\documentclass{article}
\usepackage{graphics} 
\usepackage{epsfig}
\usepackage{times}
\usepackage{amsmath}
\usepackage{amssymb}
\usepackage{threeparttable}
\usepackage{booktabs}
\usepackage{soul}
\usepackage{subfigure}
\usepackage{tabularx}
\usepackage{xcolor}
\usepackage[hidelinks]{hyperref}
\usepackage{algorithm}
\usepackage{algpseudocode}
\usepackage{comment}
\usepackage{algpseudocode}
\usepackage{float}
\usepackage{xcolor}
\usepackage{epigraph}
\usepackage{colortbl}
\definecolor{pacerTableHighlight}{HTML}{EAF3F4}
\definecolor{pacerTableSecondary}{HTML}{556676}
\floatstyle{ruled}
\restylefloat{algorithm}
\makeatletter
\algrenewcommand\algorithmiccomment[1]{\hfill{\footnotesize\textcolor{gray}{\#~#1}}}
\makeatother
\usepackage{tcolorbox}
\usepackage{multirow}
\usepackage{graphicx}
\usepackage{adjustbox}
\usepackage{censor}
\def\censorcolor{gray!50} \let\svcensorrule\censorrule \renewcommand\censorrule[1]{ \textcolor{\censorcolor}{\svcensorrule{#1}}}

\usepackage[preprint]{corl_2026}

\title{PACER: Process-Aware VLA Policy Learning from Human Corrections for Robotic Disassembly}

\author{
Chang Liu\textsuperscript{1},
Sibo Tian\textsuperscript{1},
Litian Gong\textsuperscript{2}, \\
\textbf{Zhiwen Fan}\textsuperscript{1},
\textbf{Jiachen Li}\textsuperscript{3},
\textbf{Xiao Liang}\textsuperscript{1},
\textbf{Minghui Zheng}\textsuperscript{1,*}\\
\textsuperscript{1}Texas A\&M University
\qquad
\textsuperscript{2}University of California, Riverside\\
\textsuperscript{3}Georgia Institute of Technology\\
\textsuperscript{*}Corresponding author:
\texttt{mhzheng@tamu.edu}\\~\\
\textbf{Project Page:}
\href{https://pacer-corl.github.io/}{\textcolor{blue}{\textbf{https://pacer-corl.github.io/}}}
}

\begin{document}

\maketitle

\begin{abstract}
    Robotic disassembly of end-of-life (EoL) products requires robots to accommodate product variation, follow prerequisite sequences, and approach components precisely for contact-rich manipulation. Supervised fine-tuning (SFT) provides a practical path for adaptation, but limited and costly disassembly data cannot capture the distribution shifts arising from changing environments and product uncertainties. The policy's own rollouts record its approach attempts, and human corrections demonstrate recovery from its errors, but turning these experiences into better policy performance remains challenging. We introduce PACER, a Process-Aware Correction and Evidence Reweighting framework that learns from these experiences through SFT, using progress toward the target and each trace's source and outcome to weight valid action chunks in the native supervised action loss. PACER trains one policy per weighting setting and selects one of the resulting policies through held-out validation. We evaluate PACER on UR5e motherboard disassembly across five component classes, where it achieves 87/100 successful approach-and-handoff trials on the held-out benchmark, compared with 54/100 for vanilla post-SFT.
\end{abstract}


\section{Introduction}

\epigraph{\textit{``The search for truth is more precious than its possession.''}}{--- Albert Einstein}
Electronic waste (e-waste) is among the fastest-growing end-of-life (EoL) streams worldwide, but current improper recovery processes impose substantial environmental burdens and leave considerable material value unrealized \cite{liu2027transforming, balde2024global}. Disassembly is central to recycling and remanufacturing EoL products, but it remains largely manual because workers can adapt to variations in product type and condition. However, manual disassembly is labor-intensive and hard to scale \cite{liu2026raise}. Robotic automation offers an alternative, but most existing disassembly systems are engineered for known products, with labeled parts, fixed operation sequences, and hand-designed perception and motion modules. Such systems can execute a predefined procedure, but their capability to learn from disassembly experience is limited.

\begin{figure}
    \centering
    \includegraphics[width=0.95\linewidth]{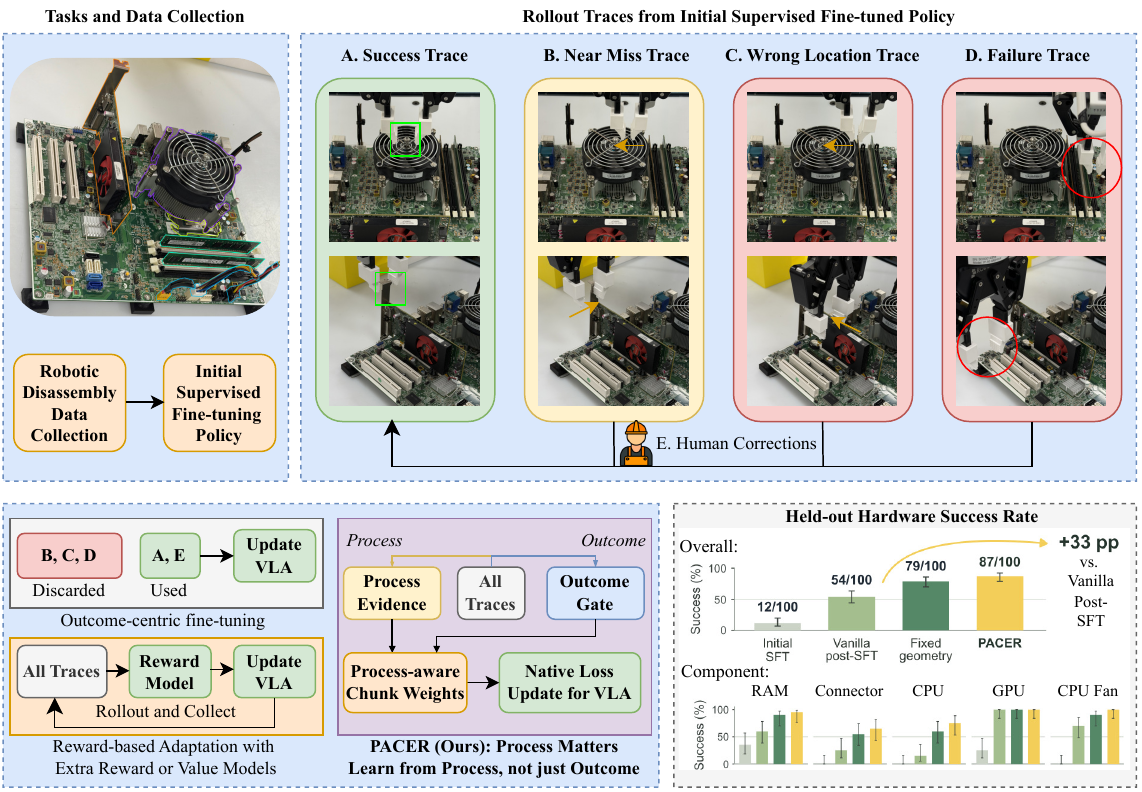}
    \caption{Compared with outcome-only filtering and reward-based adaptation, PACER uses outcome and safety annotations to determine imitation eligibility, then combines execution-process evidence with trace roles to assign bounded weights to valid action chunks in the native supervised action loss.}
    \vspace{-10pt}
    \label{fig:Intro}
\end{figure}

The rise of vision-language-action (VLA) models offers a promising direction for enabling robots to learn from operational data end-to-end \cite{kim2024openvla, team2024octo, fan2025long}. In the disassembly setting, they are used to identify actionable components in complex environments as tasks, plan motions to approach them, and execute precision contact-rich manipulations. However, these tasks differ from the daily-life tabletop tasks on which most current state-of-the-art VLA models are trained. 

Adapting VLA models to industrial disassembly requires domain post-training. Supervised fine-tuning (SFT) is the most direct approach, but its effectiveness is limited by the scarcity and narrow coverage of real disassembly demonstrations \cite{li2025transferring, liu2025vision}. While several agentic systems use VLAs primarily as planners or motion proposers rather than full executors to reduce complexity during contact-rich manipulation \cite{liu2026self, guo2026papo}, these models still suffer from distribution shift. A policy trained on nominal demonstrations can reach unfamiliar states through its own decisions during deployment. Autonomous rollouts record successful approaches, partial progress, and the states reached during failed attempts. Human corrections collected during rollouts provide recovery actions from policy-induced failure states that clean trajectories may not cover.

However, the value of robots' own rollout data and human corrections is obscured when they are simply added to fine-tuning datasets. Clean demonstrations, expert recoveries, autonomous successes, near misses, and failed rollouts differ in reliability and in how they should influence the policy update for better performance. A binary success-or-failure label is also too coarse to explain which parts of a rollout were useful, while asking humans to annotate every step in real-world rollouts is costly. Reinforcement learning (RL) offers a formal mechanism for credit assignment from trial experience to drive improvement, and recent work uses it to seek gains beyond SFT baselines in VLA post-training \cite{li2025simplevla, chen2025pi_}. However, direct RL in industrial disassembly is costly because contact-induced physical changes are difficult to simulate accurately and efficiently, and learning on real hardware is slow, costly, and safety-sensitive \cite{li2025controlvla}. Online RL obtains feedback through continued interaction, whereas offline RL learns from existing records using reward signals, value estimates, or advantage estimates \cite{zhang2026balancing, levine2020offline, peng2019advantage}. Additionally, both RL routes require task-relevant reward specification and may involve adapting the optimization method to the VLA action head \cite{chen2025pi_}. We therefore investigate whether SFT can learn more from autonomous experience and human corrections through a unified framework that retains the model's native supervised action loss.

We introduce PACER, a process-aware correction and evidence reweighting framework for VLA post-training in robotic disassembly, as shown in Figure~\ref{fig:Intro}. PACER learns from autonomous rollouts, human corrections, and supplementary clean demonstrations. It assesses progress toward the intended target within each trace and uses the trace's role to judge which recorded actions provide useful supervision. These assessments translate into bounded weights that scale each valid action chunk's contribution to the native supervised action loss. Different weighting settings specify how process evidence and trace roles are combined to determine these weights. PACER fine-tunes the same reference policy under each setting and selects one of the resulting policies through offline validation. We evaluate the selected policy on real UR5e motherboard disassembly across five component classes. PACER achieves 87/100 successful approach-and-handoff trials on the held-out benchmark, compared with 79/100 for fixed-geometry reweighting and 54/100 for vanilla post-SFT.

\section{Related Work}

\textbf{Vision-language-action models.}
Vision-language-action (VLA) models reason over multimodal inputs, including task instructions, enabling the system to use the visual state, understand language prompts, and generate executable actions for robot control in a single model \cite{sapkota2025vision, kawaharazuka2025vision, din2025vision}. Systems such as RT-2 \cite{brohan2023rt}, Octo \cite{team2024octo}, OpenVLA \cite{kim2024openvla}, OpenVLA-oft \cite{kim2025fine}, $\pi_{0.5}$ \cite{intelligence2025pi_}, and $\pi_{0.7}$ \cite{intelligence2026pi} show that visual robot policies can transfer across tasks, embodiments, and instructions when trained on large robot datasets. DROID \cite{khazatsky2024droid} provides real-world manipulation data, while LIBERO \cite{liu2023libero} benchmarks knowledge transfer in lifelong robot learning. However, most of these models are trained on and targeted at daily-life tabletop tasks under a general manipulation setting. Industrial practice, like disassembly, introduces a different regime in which the policy must operate amid product variation, limited demonstrations, contact-rich execution, and costly real-world evaluation \cite{li2025transferring, zhang2025pure}.

\begin{figure}
    \centering
    \includegraphics[width=0.98\linewidth]{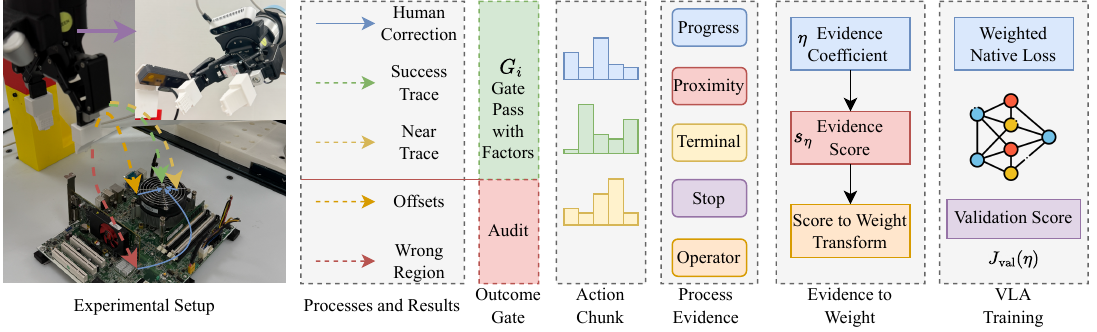}
    \caption{PACER converts autonomous rollouts, operator corrections, and supplementary clean demonstrations into a weighted native-loss training signal by checking imitation eligibility and weighting valid action chunks with process evidence.}
    \label{fig:2}
\end{figure}

\textbf{VLA post-training and learning from human corrections.} Post-training is the standard mechanism for adapting a pretrained VLA to a specific deployment domain and enhancing task performance \cite{xiang2025parallels}. Advantage-weighted regression and AWAC use exponentiated estimated advantages as nonnegative weights on a behavior-cloning loss \cite{peng2019advantage, nair2020awac}. L2RW updates minibatch example weights through validation-based meta-gradients, whereas Meta-Weight-Net uses validation feedback to learn a loss-to-weight function \cite{ren2018learning, shu2019meta}.

Reinforcement learning (RL) extends this principle from supervised reweighting to interactive credit assignment with rollouts and reward signals \cite{li2025simplevla, liu2026can}. Recent VLA models employ online policy-gradient post-training under sparse success rewards \cite{tan2025interactive, lu2025vla, zhang2025robustvla}, and extend this approach to deployment settings in which teleoperated interventions are fused with autonomous rollouts through advantage-conditioning, as exemplified by RECAP in $\pi^*_{0.6}$ and SOP \cite{intelligence2025pi, pan2026sop}. For learning from human corrections, HandITL and DexHiL embed low-latency
arm-hand intervention and prioritize corrective segments during training
\cite{han2026dexhil, li2026hand}. Action preference optimization converts intervention trajectories into preference pairs or language-grounded latent edits \cite{xia2026human}. Hi-ORS applies reward-threshold rejection sampling and reward-weighted supervision to accepted rollouts \cite{lu2025human}, and a VLM-based reflective post-training combines PPO guided by dense rewards from failure analysis with supervised fine-tuning on successful experience \cite{li2025reflection}. Flow-matching VLA heads do not directly expose tractable action log-likelihoods, complicating likelihood-based policy-gradient updates \cite{chen2025pi_}. Existing approaches enable policy optimization through stochastic denoising reformulations \cite{chen2025pi_} or perform RL in noise space with a frozen flow-matching decoder \cite{lu2026unified}. Offline and reward-verified variants can reduce the need for live rollouts but still rely on value estimates or reward supervision \cite{zhang2026balancing, chen2025conrft, li2025vla}.

\section{PACER}
We introduce PACER, a process-aware correction and evidence-reweighting framework for VLA post-training from heterogeneous disassembly traces, as shown in Figure~\ref{fig:2}. PACER operates on the data side and preserves the action representation, the VLA backbone architecture, and the native per-sample action prediction objective, modifying only the sample-level contribution of each training row to that objective. Adaptation is formulated as a validation-selected reweighting problem over typed rollouts, human corrections, and supplementary clean demonstrations collected after an initial supervised training.

Starting from an initial demonstration set $\mathcal{D}_{\mathrm{demo}}$, PACER trains a supervised reference policy $\pi_{\theta_0}$ and uses it to construct a separate post-training dataset $\mathcal{D}$. The dataset contains autonomous rollouts, operator-initiated corrections, and supplementary clean demonstrations collected under a fixed coverage protocol. Each action-chunk row is assigned a process role from its provenance and operator outcome label. Eligibility gates determine which rows can provide imitation targets. For eligible rows, process evidence and role multipliers determine their contribution to training. Each candidate weighting setting $\eta$ in a finite predeclared set $\mathcal{C}$ specifies this mapping to bounded training weights, while the evidence features, process roles, and eligibility gates remain fixed. 

Adapter learning and weight selection are separated. For each setting, the inner loop initializes a fresh low-rank adaptation (LoRA) adapter from the shared SFT checkpoint $\theta_0$ and trains it under the corresponding weighted action loss. A fixed analytic validator then scores new action chunks predicted by each fitted policy from logged validation observations. The outer loop selects the evidence-to-weight configuration over $\mathcal{C}$ by validation score, subject to predefined validity and safety checks (audits). The selected policy is then evaluated on held-out hardware tasks. Algorithm~\ref{alg:pacer} summarizes the procedure.

\begin{algorithm}[t]
\caption{PACER framework.}
\label{alg:pacer}
    \begin{algorithmic}[1]
        \Require Initial demonstrations $\mathcal{D}_{\mathrm{demo}}$, finite predeclared weighting settings $\mathcal{C} = \{\eta_1, \ldots, \eta_M\}$.
        \Ensure Selected setting $\eta^\star$, policy $\pi^\star$ shared across all components, held-out metrics on $\mathcal{T}_{\mathrm{test}}$.
        \vspace{2pt}
        \Statex \textit{Stage 1. Trace collection and partitioning.}
        \State Fit the supervised reference $\pi_{\theta_0}$ on $\mathcal{D}_{\mathrm{demo}}$.
        \State Collect autonomous rollouts and outcome labels; immediately follow eligible unsuccessful attempts with human corrections from the reached states. Add clean demonstrations under the fixed coverage protocol.
        \State Compile typed rows into $\mathcal{D}$; fix the case-group split $(\mathcal{D}_{\mathrm{tr}}, \mathcal{D}_{\mathrm{val}})$ and reserve $\mathcal{T}_{\mathrm{test}}$ before candidate fitting or scoring.
        \vspace{2pt}
        \Statex \textit{Stages 2 to 4. Per-candidate evidence weighting, training, and validation.}
        \State Compute fixed evidence $\{e_k(i)\}_{k\in\mathcal{K}}$ from recorded chunks and eligibility gates $g_i$.
        \State Initialize $\mathcal{C}_{\mathrm{audit}} \leftarrow \emptyset$.
            \For{$\eta \in \mathcal{C}$}
              \State Compute $s_\eta(i)$ and compile offline weights $w_i(\eta)$ by Eq.~\eqref{eq:weight}, using eligible training rows for group statistics.
              \State Run the pre-training audit; skip $\eta$ if any check fails.
              \State Fit a fresh LoRA adapter $\pi_{\theta(\eta)}$ from $\theta_0$ with fixed $w_i(\eta)$ and loss $\mathcal{L}_{\mathrm{PACER}}(\theta;\eta)$ on $\mathcal{D}_{\mathrm{tr}}$.
              \State Query $\pi_{\theta(\eta)}$ on logged $\mathcal{D}_{\mathrm{val}}$ observations; score predicted chunks on $\mathcal{D}_{\mathrm{val}}^{\mathrm{score}}$ to obtain $v_j(\eta)$ and the component-balanced $J_{\mathrm{val}}(\eta)$.
              \State Add $\eta$ to $\mathcal{C}_{\mathrm{audit}}$ only if its validation predictions pass the post-training audit.
            \EndFor
        \vspace{2pt}
        \Statex \textit{Stage 5. Selection and held-out hardware evaluation.}
        \State If $\mathcal{C}_{\mathrm{audit}} = \emptyset$, set $\eta^\star = \varnothing$ and $\pi^\star = \pi_{\theta_0}$. Otherwise, select $\eta^\star = \arg\max_{\eta \in \mathcal{C}_{\mathrm{audit}}} J_{\mathrm{val}}(\eta)$ and set $\pi^\star = \pi_{\theta(\eta^\star)}$.
        \State Evaluate $\pi^\star$ on $\mathcal{T}_{\mathrm{test}}$ and report overall and per-component success rates.
    \end{algorithmic}
\end{algorithm}

\textbf{Data collection.} An initial supervised reference policy $\pi_{\theta_0}$ is fitted on the demonstration set $\mathcal{D}_{\mathrm{demo}}$. The reference policy is then deployed as an action-chunk planner across the deployment scenes while the downstream execution stack is kept fixed. Following the agentic VLA framework of SELF-VLA \cite{liu2026self}, we treat a rollout as successful when the policy reaches a region from which the downstream skill can be executed reliably. After each rollout, the operator assigns one of five labels distinguishing success, near misses, invalid orientation or location, wrong regions or targets, and uncertain cases. Each eligible unsuccessful attempt is immediately followed by a teleoperated correction from the reached state. Under a coverage rule fixed before the rollout campaign, supplementary clean demonstrations are collected only for component-phase cells where the reference policy did not achieve success in the initial supervised rollout diagnostic. These rollouts, corrections, and supplementary clean demonstrations are compiled into typed action-chunk rows to form the separate post-training buffer $\mathcal{D}$.

Before any candidate weighting is computed, $\mathcal{D}$ is partitioned once into a training row set $\mathcal{D}_{\mathrm{tr}}$ and a validation row set $\mathcal{D}_{\mathrm{val}}$. A case group bundles an autonomous rollout with any paired correction and any paired supplementary clean demonstration generated from the same case. All rows compiled from a case group are assigned to the same side of the fixed split. The held-out hardware test set $\mathcal{T}_{\mathrm{test}}$ consists of scene configurations disjoint from $\mathcal{D}_{\mathrm{demo}}$ and $\mathcal{D}$ and is reserved for final real-robot evaluation after policy selection.

Each entry of $\mathcal{D}$ is a typed action-chunk row $x_i$ that stores an observation $o_i$, a horizon-$H$ action chunk $a_{i,1:H}$, a binary per-horizon mask $m_{i,1:H}$, the component and phase labels $(c_i,\phi_i)$, the declared active target $t_i^{\mathrm{act}}$, its geometry and operator annotations $q_i$, provenance metadata $z_i$, and a process role
\begin{equation}
\label{eq:process_role}
\rho_i \in \mathcal{R} = \{\mathrm{clean}, \mathrm{correction}, \mathrm{auto\_success}, \mathrm{partial}, \mathrm{failure}, \mathrm{excluded}\}.
\end{equation}
Roles are determined by trace provenance and operator outcome labels before any candidate is scored. Clean rows are compiled from supplementary teleoperated demonstrations collected under the coverage protocol, and correction rows are compiled from operator-initiated recovery trajectories from policy-induced states. For autonomous rows, \emph{success} maps to auto\_success, \emph{near\_but\_not\_accurate} maps to partial, and labels indicating wrong orientation, wrong location, wrong region, or wrong target map to failure. Rows labeled \emph{operator\_uncertain\_exclude} are assigned the excluded role. Failure and excluded rows are retained for audit. The active target $t_i^{\mathrm{act}}$ is a phase-specific handoff region with position and orientation constraints. It is fixed by the task state before scoring and serves as the reference for evaluating recorded action chunks.

\textbf{Row weighting.} An eligibility gate $g_i \in \{0,1\}$ determines which recorded rows can provide imitation targets, enforcing training-split membership, mask validity, trusted provenance, safety, and exclusion of the failure and excluded roles. For eligible rows, PACER compares process evidence within groups sharing the same component, approach phase, and process role, then uses role multipliers to scale their contributions to training.

For each row, PACER reads a vector of bounded evidence features $e_k(i)\in[0,1]$ indexed by $k \in \mathcal{K} = \{\mathrm{prog}, \mathrm{prox}, \mathrm{term}, \mathrm{dir}, \mathrm{stop}, \mathrm{op}, \mathrm{prov}\}$. The first four are geometric and describe tool-center-point (TCP) progress, final proximity, terminal-region entry, and target-directed displacement from the recorded chunk against the declared target $t_i^{\mathrm{act}}$. The last three describe stop and handoff agreement with the expected phase, the operator outcome label, and a binary provenance check. Full feature definitions are in Appendix~\ref{app:features}.

A candidate weighting setting $\eta \in \mathcal{C}$, shared across all components, specifies nonnegative evidence coefficients $\beta_k$ for each $k \in \mathcal{K}$, role multipliers $\gamma_r$ for each $r \in \mathcal{R}$, a temperature $\tau$, an upper clip $w_{\max}$ on the final row weight, and two protocol floors $f_{\mathrm{clean}}$ and $f_{\mathrm{corr}}$ for the clean and correction roles, with $0 \leq f_{\mathrm{clean}}, f_{\mathrm{corr}} \leq w_{\max}$. The scalar process score for row $i$ under setting $\eta$ is $s_\eta(i) = \sum_{k\in\mathcal{K}}\beta_k\,e_k(i)$.

PACER centers $s_\eta$ within each component-phase-role group $\mathrm{str}(i) = (c_i, \phi_i, \rho_i)$ using the median $b_{\mathrm{str}(i)}(\eta)$ and robust scale $\sigma_{\mathrm{str}(i)}(\eta)$ of eligible training rows under the same setting. The scale is based on the median absolute deviation with a fixed floor. Every group with at least one eligible training row uses its own statistics. Only empty groups back off to the component-phase level and then the global eligible training pool. The row weight is
\begin{equation}
\label{eq:weight}
w_i(\eta) \;=\; g_i \,\mathrm{clip}\!\Bigl(\max\!\bigl(f_{\rho_i},\;\alpha_\eta(\rho_i)\,\exp\!\bigl(\tfrac{s_\eta(i)-b_{\mathrm{str}(i)}(\eta)}{\tau\,\sigma_{\mathrm{str}(i)}(\eta)+\epsilon}\bigr)\bigr),\,0,\,w_{\max}\Bigr),
\end{equation}
where $\alpha_\eta(\rho_i)$ returns the role multiplier $\gamma_{\rho_i}$ from $\eta$, $\epsilon$ is a small numerical constant, and $f_{\rho_i}$ is the protocol floor for row $i$, equal to $f_{\mathrm{clean}}$ for clean rows, $f_{\mathrm{corr}}$ for correction rows, and zero otherwise. For eligible rows, the temperature controls how sharply above-median process evidence is emphasized, the role multiplier scales the exponential weight, and the floors retain clean and correction supervision. The final clip, applied after the role multiplier and floor, bounds $w_i(\eta)$ to $[0,w_{\max}]$. The eight predeclared candidates vary the progress, proximity, terminal-region, and stop-signal coefficients, the correction and partial role multipliers, the temperature, and the upper clip. All other fields are fixed by protocol. Appendix~\ref{app:candidates} gives the complete field definitions and candidate values.

\textbf{Weighted training.} Let $\ell_{\mathrm{act}}(\theta;i,h)$ denote the policy's native per-horizon action loss inherited from $\pi_{\theta_0}$. For the flow-matching VLA used in our experiments, this is the original flow regression loss, and the same form applies to any action head that exposes a per-horizon supervised target. For each minibatch $\mathcal{B}\subset\mathcal{D}_{\mathrm{tr}}$, the adapter for weighting setting $\eta$ is updated using the batch-normalized loss
\begin{equation}
\label{eq:loss}
\mathcal{L}_{\mathrm{PACER}}(\theta;\eta)\;=\;\frac{\sum_{i\in\mathcal{B}}\sum_{h=1}^{H} w_i(\eta)\,m_{i,h}\,\ell_{\mathrm{act}}(\theta;i,h)}{\max\!\left(\sum_{i\in\mathcal{B}}\sum_{h=1}^{H} w_i(\eta)\,m_{i,h},\, \epsilon\right)}.
\end{equation}
For each weighting setting, scalar row weights are computed offline from recorded evidence using training-only group statistics and remain fixed throughout the fit. Each weight $w_i(\eta)$ is shared across the row's horizons, while the binary mask $m_{i,h}$ selects valid horizons. Minibatch normalization in Eq.~\eqref{eq:loss} keeps the loss scale comparable across settings. The native action targets are unchanged, and each candidate is fitted from the shared supervised reference checkpoint $\theta_0$ under the same post-training recipe.

\textbf{Policy selection.} Each fitted policy $\pi_{\theta(\eta)}$ is scored on $\mathcal{D}_{\mathrm{val}}$ using a predeclared analytic rule. The scoring subset $\mathcal{D}_{\mathrm{val}}^{\mathrm{score}}$ contains clean, correction, rollout success, and partial validation rows. This subset is fixed before candidate scoring and restricted to planner-approach rows. The audit subset $\mathcal{D}_{\mathrm{val}}^{\mathrm{audit}}$ contains failure and excluded rows for assessing wrong-target and unsafe-behavior rates from the fitted policies' predictions.

For each scoring row $j \in \mathcal{D}_{\mathrm{val}}^{\mathrm{score}}$, the policy is queried on the logged observation, and its predicted action chunk is scored against task primitives and reference rows. The bounded metrics $\tilde e_m(j,\eta)\in[0,1]$ over the applicable set $\mathcal{M}_j$ describe predicted behavior, while the training evidence describes recorded chunks. The geometric terms measure progress, proximity, terminal entry, and direction using the same primitives as the training evidence. Other terms measure stop and handoff correctness, alignment to a matched clean or correction chunk when available, and no-regression against $\pi_{\theta_0}$. The coefficients $\{\lambda_m\}_{m\in\mathcal{M}_j}$ are nonnegative, declared before ranking, and renormalized over the terms applicable to row $j$.

The binary validation blocker $B_j(\eta)$ acts on each predicted action chunk, setting the row score to zero for wrong-target endpoints, entry into declared non-target regions, violations of the endpoint orientation tolerance, or unsafe predictions. The geometric tolerances and blocker predicates are specified in Appendix~\ref{app:components} and \ref{app:blocker}. The row-level validation score is
\begin{equation}
\label{eq:row_val}
v_j(\eta)\;=\;B_j(\eta)\,\mathrm{clip}\!\left(\frac{\sum_{m\in\mathcal{M}_j}\lambda_m\,\tilde e_m(j,\eta)}{\sum_{m\in\mathcal{M}_j}\lambda_m+\epsilon},\, 0,\, 1\right).
\end{equation}

Let $\mathcal{G}_{\mathrm{val}}$ be the set of components with scoring rows and $\mathcal{D}_{\mathrm{val}}^c$ the validation rows of component $c$. The scoring subset for component $c$ is $\mathcal{D}_{\mathrm{val}}^{\mathrm{score},c} = \mathcal{D}_{\mathrm{val}}^{\mathrm{score}} \cap \mathcal{D}_{\mathrm{val}}^c$. The component-balanced validation score is
\begin{equation}
\label{eq:j_val}
J_{\mathrm{val}}(\eta)\;=\;\frac{1}{|\mathcal{G}_{\mathrm{val}}|}\sum_{c\in\mathcal{G}_{\mathrm{val}}}\frac{1}{|\mathcal{D}_{\mathrm{val}}^{\mathrm{score},c}|}\sum_{j\in\mathcal{D}_{\mathrm{val}}^{\mathrm{score},c}} v_j(\eta).
\end{equation}
The component-balanced average gives each component equal weight in candidate ranking.

The pre-training audit checks whether a candidate's compiled weights respect the split and eligibility rules, remain bounded, and provide sufficient effective training support. The post-training audit checks the fitted policy's wrong-target and unsafe-behavior rates over $\mathcal{D}_{\mathrm{val}}^{\mathrm{score}} \cup \mathcal{D}_{\mathrm{val}}^{\mathrm{audit}}$ under the declared predicates and tolerances. It also checks per-component non-regression against $\pi_{\theta_0}$ on $\mathcal{D}_{\mathrm{val}}^{\mathrm{score}}$. Candidates that pass both audits form the ranked set $\mathcal{C}_{\mathrm{audit}}$. The full scoring metrics and audit criteria, including the effective weighted horizon count $n_{\mathrm{eff}}(\eta)$, are given in Appendix~\ref{app:valscore}--\ref{app:audits}.
When $\mathcal{C}_{\mathrm{audit}} \neq \emptyset$, PACER selects
\begin{equation}
\label{eq:selection}
\eta^\star\;=\;\arg\max_{\eta\in\mathcal{C}_{\mathrm{audit}}} J_{\mathrm{val}}(\eta), \qquad \pi^\star = \pi_{\theta(\eta^\star)}.
\end{equation}
If $\mathcal{C}_{\mathrm{audit}} = \emptyset$, PACER sets $\eta^\star = \varnothing$ and selects the supervised reference policy $\pi^\star = \pi_{\theta_0}$. For the main benchmark, we evaluate the selected policy $\pi^\star$ on the held-out hardware tasks $\mathcal{T}_{\mathrm{test}}$.

\section{Experiments}

\textbf{Setup.} We evaluate PACER on a UR5e motherboard-disassembly platform across five components: CPU, GPU, CPU fan, RAM, and the connector block. The supervised reference policy $\pi_{\theta_0}$ is obtained by supervised fine-tuning of $\pi_{0.5}$-DROID on an initial set $\mathcal{D}_{\mathrm{demo}}$ of 710 teleoperated trajectories. RAM and connector-block demonstrations are split across multiple hardware configurations, yielding lower per-configuration coverage than the three single-configuration components.

After supervised fine-tuning, $\pi_{\theta_0}$ is deployed to collect 70 autonomous rollouts per component for the post-training dataset $\mathcal{D}$. The success counts are 4, 31, 11, 7, and 2 for CPU, GPU, CPU fan, RAM, and the connector block, respectively. Eligible unsuccessful attempts receive operator corrections, and supplementary clean demonstrations are added under the coverage protocol. The split assigns 60 training and 10 validation case groups per component, with validation configurations absent from training.

All post-training policies start from $\pi_{\theta_0}$ and use the LoRA and optimization settings in Appendix~\ref{app:impl}. Training uses a batch size of 32 and 10,000 training steps for each policy. PACER trains one policy for each of eight predeclared weighting settings and selects one using $J_{\mathrm{val}}$, whereas each baseline trains one policy under a fixed setting. The held-out hardware test set $\mathcal{T}_{\mathrm{test}}$ contains two scene configurations per component that do not appear in either $\mathcal{D}_{\mathrm{demo}}$ or $\mathcal{D}$, with 10 trials per configuration. Each method is therefore evaluated on 20 trials per component and 100 trials in total. A trial succeeds when the operator judges that the fixed downstream skill can execute reliably from the reached state and the trial ends without timeout, unsafe abort, or intervention before a valid handoff state (Appendix~\ref{app:labels}). A manual terminal stop is allowed only after the policy reaches the handoff region. The metric evaluates planner approach and handoff, with scored trials ending before downstream skill execution.

We compare PACER with the supervised reference $\pi_{\theta_0}$ and six post-training baselines. Vanilla post-SFT uses supplementary clean demonstrations and corrections from $\mathcal{D}$ with uniform weights, while Uniform all-eligible additionally includes eligible autonomous traces and assigns equal weights to all included rows. Fixed geometry combines progress, proximity and target-region membership with fixed role multipliers, providing the closest fixed-rule comparison to PACER. The remaining baselines use narrower subsets of $\mathcal{D}$. Outcome only uses supplementary clean demonstrations and autonomous successes, while Clean only and Correction only use supplementary demonstrations and corrections, respectively. We specify each baseline's training data and weighting rule in Appendix~\ref{app:baselines}.

\textbf{Results.} The validation scores and hardware success counts, with marginal 95\% Wilson intervals for the overall success rates, are reported in Table~\ref{tab:main_results}. The validation score $J_{\mathrm{val}}$ is used for selection within PACER's eight-candidate family and as a diagnostic for the baselines. In this run, PACER selects the setting that emphasizes terminal-region and stop-signal evidence. The selected policy succeeds in 87 of 100 trials, compared with 79/100 for the fixed-geometry comparator, 64/100 for uniform all-eligible training and 54/100 for vanilla post-SFT. In this disassembly setting, PACER improves approach-and-handoff success by 33 percentage points over vanilla post-SFT while retaining the native action loss. Table~\ref{tab:app_diff} provides descriptive intervals for PACER's success-rate differences relative to other baselines. PACER achieves higher success counts than Fixed geometry on RAM, the connector block, CPU and CPU fan, while the two policies match on GPU. The Clean only and Correction only baselines have lower overall success than vanilla post-SFT, and the correction-only policy records no successful approaches from nominal scene initializations. The selected terminal\_stop\_heavy setting also has the highest observed success rate in a separate internal hardware comparison of all eight candidates (Appendix~\ref{app:internal8}).

\begin{table}[!]
\centering
\small
    \setlength{\tabcolsep}{3pt}
    \renewcommand{\arraystretch}{1.10}
    \caption{Validation scores and held-out hardware success counts. Each method has 10 trials on each of two configurations per component, one close to the training distribution and one out of distribution. Brackets give marginal Wilson 95\% confidence intervals for the overall success rate, in percent. Bold marks success-count column maxima.}
    \label{tab:main_results}
    \fontencoding{T1}\selectfont
    \begin{tabularx}{\linewidth}{@{}>{\raggedright\arraybackslash}p{0.25\linewidth}>{\raggedleft\arraybackslash}X>{\color{pacerTableSecondary}}r*{4}{>{\raggedleft\arraybackslash}X}rr@{}}
    \toprule
    & \multicolumn{2}{c}{Overall ($n=100$)} & \multicolumn{5}{c}{Per component ($n=20$)} & Offline \\
    \cmidrule(lr){2-3}\cmidrule(lr){4-8}\cmidrule(l){9-9}
    Method & Success & \multicolumn{1}{r}{95\% CI (\%)} & RAM & Conn. & CPU & GPU & CPU fan & $J_{\mathrm{val}}$ \\
    \midrule
    Initial SFT $\pi_{\theta_0}$ & 12 & $[7.0, 19.8]$ & 7 & 0 & 0 & 5 & 0 & $0.7214$ \\
    \midrule
    Vanilla post-SFT & 54 & $[44.3, 63.4]$ & 12 & 5 & 3 & \textbf{20} & 14 & $0.7579$ \\
    Uniform all-eligible & 64 & $[54.2, 72.7]$ & 16 & 3 & 8 & \textbf{20} & 17 & $0.7648$ \\
    Outcome only & 11 & $[6.3, 18.6]$ & 0 & 0 & 0 & 10 & 1 & $0.7340$ \\
    Clean only & 24 & $[16.7, 33.2]$ & 0 & 0 & 9 & 8 & 7 & $0.7425$ \\
    Correction only & 0 & $[0.0, 3.7]$ & 0 & 0 & 0 & 0 & 0 & $0.7201$ \\
    Fixed geometry & 79 & $[70.0, 85.8]$ & 18 & 11 & 12 & \textbf{20} & 18 & $0.7699$ \\
    \rowcolor{pacerTableHighlight}
    \textbf{PACER selected} & \textbf{87} & $[79.0, 92.2]$ & \textbf{19} & \textbf{13} & \textbf{15} & \textbf{20} & \textbf{20} & $0.7707$ \\
    \bottomrule
\end{tabularx}
\end{table}

\begin{table}[!]
\centering
\small
\setlength{\tabcolsep}{3pt}
\renewcommand{\arraystretch}{1.10}
\caption{Controlled OOD comparison. Each policy has 50 trials on each of two OOD configurations, with mild and strong shifts. Overall intervals are Wilson 95\% CIs in percent. Differences are selected PACER minus each row with paired bootstrap 95\% CIs. Bold marks success-count column maxima.}
\label{tab:controlled_rebuttal}
\fontencoding{T1}\selectfont
\begin{tabularx}{\linewidth}{@{}>{\raggedright\arraybackslash}p{0.32\linewidth}*{3}{>{\raggedleft\arraybackslash}X}>{\color{pacerTableSecondary}}rr>{\color{pacerTableSecondary}}r@{}}
\toprule
& \multicolumn{3}{c}{Success counts} & & \multicolumn{2}{c}{PACER gap} \\
\cmidrule(lr){2-4}\cmidrule(l){6-7}
Policy & Mild & Strong & Overall & \multicolumn{1}{r}{95\% CI (\%)} & $\Delta$ (pp) & \multicolumn{1}{r}{95\% CI} \\
& $n=50$ & $n=50$ & $n=100$ & & & \\
\midrule
\rowcolor{pacerTableHighlight}
\textbf{PACER selected} & \textbf{43} & \textbf{29} & \textbf{72} & $[62.5, 79.9]$ & Reference & \textcolor{black}{---} \\
\addlinespace[4pt]
\multicolumn{7}{@{}l@{}}{\textcolor{pacerTableSecondary}{\textit{Fixed weighting rules}}} \\
Fixed geometry & 35 & 16 & 51 & $[41.3, 60.6]$ & $+21$ & $[13, 29]$ \\
RW-like & 33 & 18 & 51 & $[41.3, 60.6]$ & $+21$ & $[12, 30]$ \\
Uniform & 37 & 11 & 48 & $[38.5, 57.7]$ & $+24$ & $[16, 32]$ \\
Role only & 30 & 9 & 39 & $[30.0, 48.8]$ & $+33$ & $[24, 42]$ \\
\addlinespace[4pt]
\multicolumn{7}{@{}l@{}}{\textcolor{pacerTableSecondary}{\textit{Data-source removals (selected setting fixed)}}} \\
No clean & 34 & 1 & 35 & $[26.4, 44.7]$ & $+37$ & $[30, 44]$ \\
No correction & 13 & 0 & 13 & $[7.8, 21.0]$ & $+59$ & $[52, 66]$ \\
No outcome & 36 & 14 & 50 & $[40.4, 59.6]$ & $+22$ & $[14, 30]$ \\
\bottomrule
\end{tabularx}
\end{table}

\section{Ablation Study}

To validate the effectiveness of PACER, we assess the importance of supplementary clean demonstrations, human corrections, and autonomous rollouts for PACER post-training by removing each source in turn. We also compare selected PACER with four fixed weighting alternatives, as shown in Table~\ref{tab:controlled_rebuttal}. Each of the eight policies in this comparison is trained from $\pi_{\theta_0}$ with the same nominal training budget. Evaluation uses two out-of-distribution (OOD) configurations with mild and strong shifts, distinct from those in $\mathcal{T}_{\mathrm{test}}$. There are 50 trials per policy per configuration and 800 trials in total. Trials are paired across policies by component and trial index under a common reset procedure.

For the three data-source removals, new policies are trained with the selected weighting setting held fixed. No clean removes supplementary clean demonstrations, No correction removes human corrections, and No outcome removes autonomous rollout data while retaining supplementary clean demonstrations and corrections. Fixed geometry directly uses a fixed geometric scoring rule with role multipliers, whereas RW-like uses PACER's weighting formula with one preset parameter configuration (Appendix~\ref{app:candidates}) and no validation-based selection. Uniform still assigns equal weights to usable rows, while Role only assigns weights based solely on each row's process role. 

Removing any post-training data source reduces success, with the largest decline when corrections are removed. The correction-only policy in the main benchmark records no successful approaches from nominal starts. Corrections demonstrate recovery and refinement from policy-induced states, while supplementary clean demonstrations show normal progress toward the target from nominal starts. Even with both forms of human supervision retained, the No outcome policy achieves lower success than selected PACER when autonomous rollouts are removed. This gap indicates that the policy's own execution experience adds value beyond the clean and corrective demonstrations. Among policies trained with all three data sources, selected PACER achieves higher success than the fixed weighting alternatives. The paired bootstrap interval for its success-rate difference relative to Fixed geometry lies entirely above zero. Relative to Uniform, RW-like yields only a small numerical gain, whereas Role only performs worse, indicating that the tested role weights alone are insufficient to improve learning over equal weighting.

\section{Limitations}
PACER relies on accurate trace metadata, including target geometry, process roles, operator labels, masks, and provenance fields. The geometry-sensitivity experiment examines retraining under one target-region perturbation (Appendix~\ref{app:rebuttal_sensitivity}). Robustness to a broader range of metadata errors and to corrupted training labels remains untested. Validation resampling leaves the leading candidates in the $J_{\mathrm{val}}$ ranking difficult to distinguish (Appendix~\ref{app:audits}), while calibration of this score against closed-loop hardware performance remains limited. The configuration comparisons and data-source ablations do not isolate the individual effects of evidence features, role multipliers, protocol floors, or candidate selection. The evaluation is limited to motherboard disassembly on a single robot platform and a single VLA model. PACER has not yet been evaluated in other task domains or on standard robotic manipulation benchmarks. Hardware evaluation results have not yet been used to guide iterative refinement of the PACER framework.

\section{Conclusion}
PACER enables a supervised VLA to learn from its own autonomous rollouts, human corrections, and supplementary clean demonstrations. It uses fixed process evidence to weight valid action chunks under the native supervised action loss and selects a fitted policy through offline validation. On the evaluated motherboard disassembly tasks, PACER improves approach-and-handoff success over vanilla SFT post-training and fixed-geometry reweighting. These results suggest that process-aware use of heterogeneous experience can make SFT more effective when disassembly demonstrations are limited.

\clearpage

\acknowledgments{This work was partially supported by the U.S. National Science Foundation under Grant Nos. 2422826 and 2527316. Portions of this research were conducted with the advanced computing resources provided by Texas A\&M High Performance Research Computing. We also appreciate all reviewers and chairs for their great comments to help us improve the work.}

\bibliography{Ref}  

\clearpage

\newpage

\appendix

\section{Task, Hardware, and Evaluation Protocol}
\label{app:task}

\subsection{Trial accounting and configuration terminology}
\label{app:accounting}
The main held-out hardware benchmark uses two scene configurations per component, one close to the training distribution and one out of distribution, with 10 trials per configuration. Each method is therefore evaluated on 100 held-out trials across the five components, or 20 trials per component.

The test configurations are fixed before candidate selection and do not occur in either $\mathcal{D}_{\mathrm{demo}}$ or the post-training buffer $\mathcal{D}$. The rollout and correction collection configurations are described in Section~\ref{app:rollouts}.

\subsection{Hardware platform and success criterion}
\label{app:platform}
The platform is a UR5e arm operating on motherboard disassembly tasks over five reuse-relevant components, namely CPU, GPU, CPU fan, RAM, and the connector block. Observations come from an Intel RealSense D435i wrist camera and an OAK-D Pro fixed-focus side camera. The end effector is a Robotiq 2F-85 gripper fitted with a custom extension designed for desktop disassembly tasks. The control and inference rates, action representation, coordinate frame, and action-chunk horizon $H$ are listed with the training configuration in Section~\ref{app:impl}.

The learned policy is evaluated as an action-chunk planner, and the downstream execution stack is fixed across all methods. Scored trials stop before downstream skill execution. The operator uses the final tool-center-point pose and operator-visible state to judge whether that skill can execute reliably, applying the same handoff rubric and termination criteria to every method (Section~\ref{app:labels}). The numerical tolerances in Section~\ref{app:components} are used for training evidence and offline validation.

\subsection{Component tasks, targets, and thresholds}
\label{app:components}
Each component has a declared active target and one or more approach phases. The phase $\phi$ indexes the stage of the approach task, and the scored validation rows use the planner approach and handoff phase. For each component, the positional geometry of the declared target region is given by an operator-captured point set recorded with the teleoperation joystick before any scoring. We use a shared orientation tolerance of 0.35 rad and a component-dependent position tolerance, with $r_{\mathrm{target}} = 0.005$ m for CPU and $r_{\mathrm{target}} = 0.010$ m for the other four components. The proximity normalizer is set to twice this hard tolerance, $d_{\mathrm{ref}}(c) = 2\, r_{\mathrm{target}}(c)$, giving $d_{\mathrm{ref}} = 0.010$ m for CPU and $d_{\mathrm{ref}} = 0.020$ m otherwise. Proximity evidence is therefore graded inside a narrow neighborhood of the declared target, while terminal-region membership is governed by the stricter component-specific tolerance.

Table~\ref{tab:app_prompts} gives the canonical language prompt of each component. The prompts route low-value components, high-value components, and components requiring further disassembly to the yellow, blue, and green zones. The same prompt dictionary is used for every recording and evaluation. The prompt carries the full high-level instruction for the policy, while the hardware metric scores the approach and handoff phases within that instruction.

\begin{table}[t]
\centering
\small
\caption{Canonical language prompts by component.}
\label{tab:app_prompts}
\begin{tabular}{@{}lp{0.74\linewidth}@{}}
\toprule
Component & Canonical prompt \\
\midrule
CPU & Extract the CPU by unlocking the bracket first, then pick up the CPU and place it inside the blue area as a high-value component. \\
RAM & Extract the closest visible RAM stick and place it inside the blue area as a high-value component. \\
CPU fan & Extract the CPU fan and place it inside the yellow area as a low-value component. \\
Connector & Extract the closest visible connector and place it inside the yellow area as a low-value component. \\
GPU & Extract the graphics card and place it inside the green area, as it needs subsequent disassembly. \\
\bottomrule
\end{tabular}
\end{table}

\subsection{Closed-vocabulary outcome labels and labeling protocol}
\label{app:labels}
Every autonomous rollout and every held-out evaluation trial receives exactly one of five closed-vocabulary labels, used consistently for both data collection and hardware evaluation. The label set is \emph{success}, \emph{near\_but\_not\_accurate}, \emph{wrong\_orientation\_or\_wrong\_location}, \emph{totally\_off\_wrong\_region\_or\_target}, and \emph{operator\_uncertain\_exclude}. A trial counts as a strict success only when the label is \emph{success} and the rollout did not terminate by timeout, unsafe abort, or intervention before a valid handoff state.
\begin{equation}
\label{eq:app_strict}
\mathrm{success}_{\mathrm{strict}} = \mathbf{1}\{y = \mathrm{success}\}\cdot \mathbf{1}\{\mathrm{stop\ source} \notin \mathcal{S}_{\mathrm{excl}}\},
\end{equation}
where $\mathcal{S}_{\mathrm{excl}}$ contains timeout, unsafe-abort, and intervention stop sources. A manual terminal stop is an operator stop after the policy has reached the handoff region. An intervention stop is an operator stop before a valid handoff state.

The label \emph{near\_but\_not\_accurate} marks a rollout that reaches the neighborhood of the declared target without entering the actionable region. The label \emph{wrong\_orientation\_or\_wrong\_location} covers trajectories near the intended component but with an invalid pose or local target, while \emph{totally\_off\_wrong\_region\_or\_target} covers trajectories directed to the wrong region or component. The \emph{operator\_uncertain\_exclude} label marks a rollout that the operator cannot judge confidently.

Each trial is labeled by one operator from the recorded final state using the shared closed vocabulary and rubric. Method identity is visible during execution. Labels are fixed before aggregate method performance is computed.

A borderline case is assigned to \emph{near\_but\_not\_accurate} unless the operator judges that the downstream skill can be executed reliably from the reached pose. A failure label requires a clear deviation from the declared target, the local pose, or the handoff criterion. The rollout timeout is 600 steps at 10 Hz, or 60 seconds of continuous execution. No unsafe abort occurred in the rollout campaign or the held-out evaluation.

\subsection{Held-out hardware test protocol}
\label{app:eval}
Hardware evaluation follows model selection, using the held-out configurations and trial counts in Section~\ref{app:accounting}.

A rollout ends on the model stop token, a manual stop, a fixed timeout, or an unsafe abort, after which the operator assigns one of the five labels. Each method is evaluated in a separate pass, with pass order randomized across the held-out configurations. Within each pass, components are evaluated in the fixed order: CPU fan, RAM, connector block, GPU, and CPU. The board is reset to the specified configuration before each trial.

\section{Data Collection and Splits}
\label{app:data}

\subsection{Terminology}
\label{app:glossary}
The protocol distinguishes several related units. A component is one of the CPU, GPU, CPU fan, RAM, and the connector block. A phase $\phi$ is a stage of a component approach task, and a component-phase cell is a component paired with a stage.

A post-training collection configuration is a physical placement used to collect autonomous rollouts and correction trajectories for buffer construction and the train-validation split. A case group is an indexed component-configuration collection slot used for splitting. It contains one autonomous rollout, its immediately following correction when the unsuccessful attempt is eligible, and any supplementary clean traces assigned to that slot. A held-out scene configuration is a final-evaluation setup disjoint from the collection configurations.

\subsection{Initial demonstrations}
\label{app:collection}
The reference policy $\pi_{\theta_0}$ is obtained by supervised fine-tuning of the $\pi_{0.5}$-DROID model on $\mathcal{D}_{\mathrm{demo}}$, a set of 710 teleoperated trajectories, namely 130 each for CPU, GPU, and CPU fan and 160 each for RAM and the connector block. RAM and the connector block are recorded across multiple hardware configurations, whereas CPU, GPU, and CPU fan are each recorded in a single initial demonstration configuration.

\subsection{Rollouts and Corrections}
\label{app:rollouts}
After this initial supervised fine-tuning stage, $\pi_{\theta_0}$ is deployed in seven post-training collection configurations per component. Collection in a configuration proceeds in the fixed component order: CPU fan, RAM, connector block, GPU, and CPU. The operator captures the target point set of a component with the teleoperation joystick immediately before that component's collection, and the completed component is removed before the next component begins. These point sets define the declared target regions of Section~\ref{app:components} and are used only for training evidence and offline validation scoring. In each collection configuration, the robot executes the given component task, and we record 10 rollouts, yielding 70 rollouts per component and 350 in total. Table~\ref{tab:app_dataset_stats} reports the outcomes of these autonomous rollouts.

\begin{figure}[t]
\centering
\includegraphics[width=\linewidth]{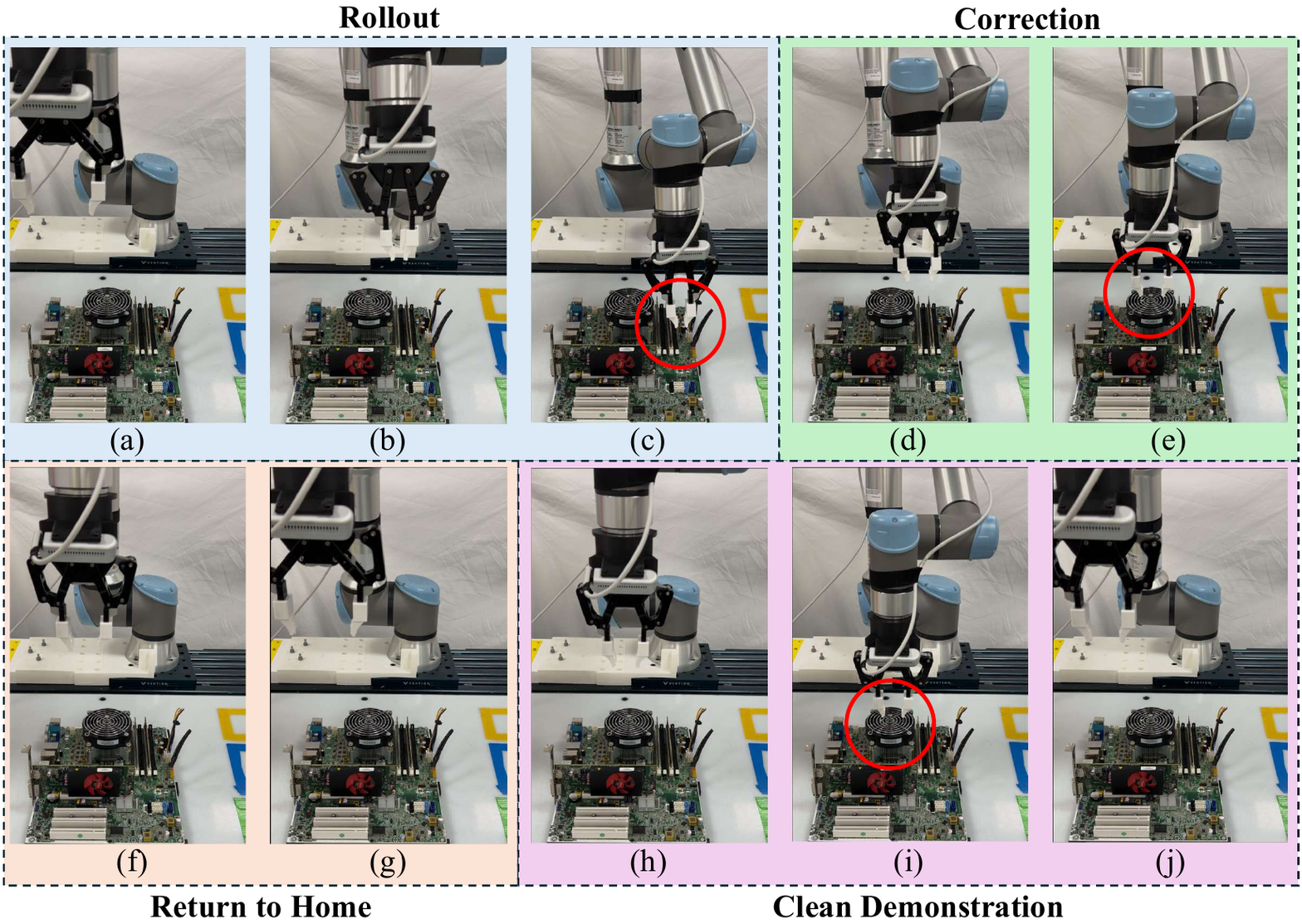}
\caption{One post-training collection cycle, starting from the initial policy after supervised fine-tuning. A rollout ends in a non-success state (a--c), an operator correction is seeded from that state (d--e), the arm returns to home (f--g), and a supplementary clean demonstration is recorded for the same cell (h--j).}
\label{fig:app_collection_loop}
\end{figure}

Each autonomous attempt ending in an eligible failure or near-miss state is immediately followed by a teleoperated correction from that state in the same collection configuration. Eligible outcomes are the near-miss and two failure labels defined in Section~\ref{app:labels}. Unsafe aborts and operator-uncertain cases are excluded. Each correction is stored as a separate trajectory while retaining its association with the failed attempt. Each component contributes at most 70 correction trajectories across its seven collection configurations. The trajectory count follows the eligible non-success outcomes; Table~\ref{tab:app_buffer} reports the corresponding compiled correction-row counts. Figure~\ref{fig:app_collection_loop} shows one collection cycle: an unsuccessful rollout, a correction from the reached state, a return to home, and a supplementary clean demonstration for the same cell.

\subsection{Supplementary clean demonstrations and disjointness}
\label{app:suppclean}
Supplementary clean demonstrations are recorded only for component-phase cells where the reference policy fails to achieve success, as observed in the initial supervised rollout diagnostic. This coverage rule is fixed before the rollout campaign; Table~\ref{tab:app_buffer} reports the resulting per-component supplementary clean row counts. An unpaired supplementary clean demonstration is assigned to the split of its collection configuration, and the case-group split applies only to the post-training buffer, which is disjoint from $\mathcal{D}_{\mathrm{demo}}$.

\subsection{Post-training buffer composition}
\label{app:datastats}
Table~\ref{tab:app_dataset_stats} reports the outcome breakdown of the 70 autonomous rollouts per component. GPU has the most successful rollouts (31/70), while the other components have 2--11 successes each. Table~\ref{tab:app_buffer} instead reports compiled training-row counts from autonomous rollouts, operator corrections, and supplementary clean demonstrations; PACER reweights these action-chunk rows. The gap between the compiled train rows and the positively weighted rows in Table~\ref{tab:app_baselines} contains the failure and excluded rows, along with positive-role rows removed by the mask, provenance, and safety factors.

\begin{table}[t]
\centering
\small
\caption{Per-component outcome distribution of the 70 autonomous rollouts across the seven collection configurations, with the failure count split into its two labels.}
\label{tab:app_dataset_stats}
\begin{tabular*}{\linewidth}{@{\extracolsep{\fill}}lccccccc@{}}
\toprule
Component & Configs & Rollouts & Success & Partial & \shortstack{Wrong\\orientation\\or location} & \shortstack{Wrong\\region\\or target} & Excluded \\
\midrule
RAM       & 7 & 70 & 7  & 3  & 6  & 54 & 0 \\
Connector & 7 & 70 & 2  & 25 & 23 & 20 & 0 \\
CPU       & 7 & 70 & 4  & 8  & 6  & 52 & 0 \\
GPU       & 7 & 70 & 31 & 37 & 2  & 0  & 0 \\
CPU fan   & 7 & 70 & 11 & 6  & 12 & 41 & 0 \\
\bottomrule
\end{tabular*}
\end{table}

\begin{table}[t]
\centering
\small
\caption{Post-training buffer composition as compiled training-row counts after action-chunk compilation. The compiled column counts all rows in the training split, including failure and excluded rows retained for audit, and positively weighted rows per method are reported in Table~\ref{tab:app_baselines}.}
\label{tab:app_buffer}
\begin{tabular*}{\linewidth}{@{\extracolsep{\fill}}lccc@{}}
\toprule
Component & Correction rows & Supp.\ clean rows & All compiled train rows \\
\midrule
RAM       & 301 & 480 & 1616 \\
Connector & 351 & 572 & 2019 \\
CPU       & 348 & 587 & 1880 \\
GPU       & 107 & 334 & 1698 \\
CPU fan   & 269 & 364 & 1623 \\
Total     & 1376 & 2337 & 8836 \\
\bottomrule
\end{tabular*}
\end{table}

\subsection{Case-group split}
\label{app:split}
Validation is configuration-disjoint from training. For each component, the seven post-training collection configurations are split into six training configurations and one validation configuration under a fixed seed that selects the validation configuration. Every trace collected under a configuration, namely its autonomous rollouts, its correction trajectories, and any supplementary clean demonstrations, is assigned as a whole to the corresponding side. Since each component-configuration cell contributes 10 autonomous rollouts, the split yields 60 training case groups and 10 validation case groups per component.

The compiled manifest contains 8836 training, 1328 validation, and 1132 frozen rows. Frozen rows are reserved for audit and diagnostics. Table~\ref{tab:app_valdist} gives the validation counts by component and role. The validation configurations contain no auto\_success rows.

\begin{table}[t]
\centering
\small
\caption{Compiled validation rows by component and role. Clean, correction, auto\_success, and partial rows form the positive-scoring set $\mathcal{V}_c^+$, failure or excluded rows are audit-only, and the ranking pass of Table~\ref{tab:app_ranking} scores a fixed planner-phase subset of these rows.}
\label{tab:app_valdist}
\begin{tabular}{lccccc}
\toprule
Component & Clean & Correction & Auto success & Partial & Failure / excluded \\
\midrule
RAM       & 73 & 46 & 0 & 73 & 29 \\
Connector & 78 & 73 & 0 & 22 & 72 \\
CPU       & 85 & 45 & 0 & 12 & 115 \\
GPU       & 116 & 69 & 0 & 50 & 176 \\
CPU fan   & 45 & 52 & 0 & 11 & 86 \\
\bottomrule
\end{tabular}
\end{table}

\section{Evidence, Gates, and Weights}
\label{app:method}

\subsection{Evidence feature definitions}
\label{app:features}
Geometric evidence uses the full recorded action-chunk prefix through the last unmasked horizon, $h_i^{\mathrm{end}} = \max\{h : m_{i,h} = 1\}$. Rows with no valid mask mass receive zero training weight and are excluded from geometric statistics. Let $\mathbf{p}_{i,t}$ and $R_{i,t}$ denote the recorded TCP position and orientation at step $t$ in the robot base frame, and let $\mathbf{p}_i^\star$ denote the calibrated center or canonical contact point of the declared target region, fixed before scoring. Define
\begin{equation}
\label{eq:app_dist}
\begin{aligned}
d_0(i) &= \lVert \mathbf{p}_{i,0} - \mathbf{p}_i^\star \rVert, &\quad
d_{\mathrm{end}}(i) &= \lVert \mathbf{p}_{i,h_i^{\mathrm{end}}} - \mathbf{p}_i^\star \rVert, \\
d_{\min}(i) &= \min_{0 \le t \le h_i^{\mathrm{end}}} \lVert \mathbf{p}_{i,t} - \mathbf{p}_i^\star \rVert, &\quad
\Delta \mathbf{p}_i &= \mathbf{p}_{i,h_i^{\mathrm{end}}} - \mathbf{p}_{i,0}.
\end{aligned}
\end{equation}
Progress and proximity use this representative point. Terminal membership and wrong-target checks use the full declared region. Let $T_i \in \{0,1\}$ indicate target-metadata consistency and define
\begin{equation}
\label{eq:app_term}
Z_i^{\mathrm{term}} = \mathbf{1}\bigl\{(\mathbf{p}_{i,h_i^{\mathrm{end}}}, R_{i,h_i^{\mathrm{end}}}) \in \mathcal{R}^{\mathrm{term}}_{c_i, \phi_i}\bigr\},
\end{equation}
which reduces to $d_{\mathrm{end}}(i) \le r_{\mathrm{target}}$ for a position-only spherical region. The four geometric features are
\begin{equation}
\label{eq:app_geom}
\begin{aligned}
e_{\mathrm{prog}}(i) &= T_i\,\mathrm{clip}\!\left(\frac{d_0(i) - d_{\min}(i)}{d_0(i) + \epsilon}, 0, 1\right), &
e_{\mathrm{prox}}(i) &= T_i\,\mathrm{clip}\!\left(1 - \frac{d_{\mathrm{end}}(i)}{d_{\mathrm{ref}}}, 0, 1\right), \\
e_{\mathrm{term}}(i) &= T_i\, Z_i^{\mathrm{term}}, &
e_{\mathrm{dir}}(i) &= T_i\,\mathrm{clip}\!\left(\frac{\Delta \mathbf{p}_i^\top (\mathbf{p}_i^\star - \mathbf{p}_{i,0})}{\lVert \Delta \mathbf{p}_i \rVert\,\lVert \mathbf{p}_i^\star - \mathbf{p}_{i,0} \rVert + \epsilon}, 0, 1\right),
\end{aligned}
\end{equation}
Here $\mathcal{R}^{\mathrm{term}}_{c,\phi}$ is the terminal pose region for the declared active target, defined by the configuration-specific target point set and orientation constraint with the tolerances in Section~\ref{app:components}; $d_{\mathrm{ref}}(c) = 2\,r_{\mathrm{target}}(c)$. Progress uses $d_{\min}$ to retain credit for partial approaches. Missing or inconsistent component, phase, target, or target-geometry metadata sets $T_i = 0$.

The stop feature $e_{\mathrm{stop}}(i) = S_i$ is one only for a phase-ending row with the expected model stop or handoff event inside the terminal region. Otherwise it is zero.

Operator evidence $e_{\mathrm{op}}(i)$ is $1$ for clean and correction rows, $0.8$ for auto\_success, and $0.6$ for partial rows. Failure and excluded rows are role-gated out. This evidence is constant within each eligible component-phase-role group, so centering removes its additive score contribution, while the labels still determine roles and imitation eligibility. Provenance evidence $e_{\mathrm{prov}}(i) = P_i$ equals one when the manifest, timing alignment, component identity, and action mask are valid. Under the fixed coefficients in Section~\ref{app:candidates}, provenance acts through the gate and direction contributes only to validation.

\subsection{Per-horizon mask}
\label{app:mask}
The binary mask $m_{i,h}$ is one only for a recorded horizon with a finite normalized action target within calibrated bounds, a synchronized observation and proprioceptive state, and no preceding unsafe event or manual stop. All other horizons are masked out. A manual stop used as a safety intervention also sets $G_i^{\mathrm{safe}} = 0$, and $G_i^{\mathrm{mask}} = 1$ requires at least one valid horizon.

\subsection{Eligibility gate}
\label{app:gate}
The eligibility gate is the product of binary factors
\begin{equation}
\label{eq:app_gate}
g_i = G_i^{\mathrm{split}}\, G_i^{\mathrm{mask}}\, G_i^{\mathrm{prov}}\, G_i^{\mathrm{safe}}\, G_i^{\mathrm{role}},
\end{equation}
requiring training-split membership, at least one valid horizon, trusted provenance, safety, and a role other than failure or excluded. Target-metadata consistency affects geometric evidence through $T_i$ in Eq.~\eqref{eq:app_geom}.

\subsection{Robust centering and fallback}
\label{app:center}
Within each component-phase-role group, the normalization statistics use only gate-eligible training rows:
\begin{equation}
\label{eq:app_center}
\begin{aligned}
b_{\mathrm{str}}(\eta) &= \mathrm{median}\{s_\eta(i) : g_i = 1,\, \mathrm{str}(i) = \mathrm{str}\}, \\
\sigma_{\mathrm{str}}(\eta) &= \max\!\bigl(1.4826\,\mathrm{MAD}\{s_\eta(i) : g_i = 1,\, \mathrm{str}(i) = \mathrm{str}\},\, \sigma_{\min}\bigr),
\end{aligned}
\end{equation}
Here $\mathrm{MAD}$ is the median absolute deviation and $\sigma_{\min} = 10^{-6}$. Every nonempty group uses its own statistics. Empty groups back off to the component-phase pool and then the global eligible training pool. Equation~\eqref{eq:weight} converts the centered scores to weights with $\epsilon = 10^{-6}$.

\subsection{Candidate weighting settings and fixed values}
\label{app:candidates}
In this paper, the candidate search varies eight fields
\begin{equation}
\label{eq:eta_free}
\eta_{\mathrm{free}} = \bigl(\beta_{\mathrm{prog}}, \beta_{\mathrm{prox}}, \beta_{\mathrm{term}}, \beta_{\mathrm{stop}}, \gamma_{\mathrm{correction}}, \gamma_{\mathrm{partial}}, \tau, w_{\max}\bigr),
\end{equation}
while the remaining nine fields are fixed by protocol,
\begin{equation}
\label{eq:eta_fixed}
\eta_{\mathrm{fixed}} = \bigl(\beta_{\mathrm{dir}}, \beta_{\mathrm{op}}, \beta_{\mathrm{prov}}, \gamma_{\mathrm{clean}}, \gamma_{\mathrm{auto\_success}}, \gamma_{\mathrm{failure}}, \gamma_{\mathrm{excluded}}, f_{\mathrm{clean}}, f_{\mathrm{corr}}\bigr),
\end{equation}
with $\beta_{\mathrm{dir}} = \beta_{\mathrm{prov}} = 0$, $\gamma_{\mathrm{failure}} = \gamma_{\mathrm{excluded}} = 0$, $\beta_{\mathrm{op}} = 0.05$, $\gamma_{\mathrm{clean}} = \gamma_{\mathrm{auto\_success}} = 1.00$, $f_{\mathrm{clean}} = 0.80$, and $f_{\mathrm{corr}} = 0.65$.

Table~\ref{tab:app_candidates} lists the complete finite search space of eight predeclared settings for Eq.~\eqref{eq:eta_free}. The ram\_connector\_recovery setting is defined from pre-rollout demonstration coverage.

\begin{table}[t]
\centering
\footnotesize
\caption{The eight predeclared weighting settings. The selected candidate emphasizes terminal-region and stop-signal evidence.}
\label{tab:app_candidates}
\begin{tabular*}{\linewidth}{@{\extracolsep{\fill}}lcccccccc@{}}
\toprule
Candidate & $\beta_{\mathrm{prog}}$ & $\beta_{\mathrm{prox}}$ & $\beta_{\mathrm{term}}$ & $\beta_{\mathrm{stop}}$ & $\gamma_{\mathrm{corr}}$ & $\gamma_{\mathrm{part}}$ & $\tau$ & $w_{\max}$ \\
\midrule
rw\_like                 & 0.30 & 0.45 & 0.15 & 0.05 & 1.20 & 0.60 & 1.00 & 4.50 \\
progress\_heavy          & 0.55 & 0.20 & 0.15 & 0.05 & 1.20 & 0.60 & 1.00 & 4.50 \\
hover\_proximity\_heavy  & 0.20 & 0.60 & 0.10 & 0.05 & 1.20 & 0.60 & 1.00 & 4.50 \\
terminal\_stop\_heavy    & 0.20 & 0.25 & 0.35 & 0.15 & 1.20 & 0.60 & 1.00 & 4.50 \\
correction\_heavy        & 0.30 & 0.45 & 0.15 & 0.05 & 1.80 & 0.70 & 1.00 & 4.50 \\
conservative             & 0.30 & 0.45 & 0.15 & 0.05 & 1.20 & 0.40 & 1.50 & 2.50 \\
ram\_connector\_recovery & 0.45 & 0.30 & 0.10 & 0.05 & 1.70 & 0.90 & 1.00 & 5.00 \\
balanced\_low\_clip      & 0.30 & 0.45 & 0.15 & 0.05 & 1.20 & 0.60 & 0.85 & 3.50 \\
\bottomrule
\end{tabular*}
\end{table}

\section{Validation, Audits, and Selection}
\label{app:validation}

\subsection{Offline validation scoring}
\label{app:valscore}
The row and component-balanced scores follow Eqs.~\eqref{eq:row_val} and \eqref{eq:j_val}. Let $\mathcal{V}_c^+$ denote the compiled validation rows of component $c$ with clean, correction, auto\_success, or partial roles. The planner-phase subset $\mathcal{D}_{\mathrm{val}}^{\mathrm{score},c} \subset \mathcal{V}_c^+$ is fixed before scoring and shared across candidates. Across components it contains 294 of the 850 compiled positive rows and is used for every reported $J_{\mathrm{val}}$. Failure and excluded validation rows are audit-only.

Each fitted policy is queried once per scoring row on the logged observation using deterministic inference. Predicted chunks are inverse-normalized, interpreted open-loop from the logged start state with the hardware action convention, and mapped by robot kinematics to TCP poses in the base frame.

Geometric submetrics apply Eqs.~\eqref{eq:app_dist} to \eqref{eq:app_geom} to predicted rather than recorded chunks. The stop submetric is included only for phase-ending rows and equals one for the expected stop or handoff event, and zero otherwise. When a matched clean or correction reference chunk $r(j)$ exists, direction alignment is
\begin{equation}
\label{eq:app_align}
\tilde e_{\mathrm{align}}(j, \eta) = \mathrm{clip}\!\left(\frac{\mathbf{u}_j(\eta)^\top \mathbf{u}^{\mathrm{ref}}_{r(j)}}{\lVert \mathbf{u}_j(\eta) \rVert\,\lVert \mathbf{u}^{\mathrm{ref}}_{r(j)} \rVert + \epsilon}, 0, 1\right),
\end{equation}
where $\mathbf{u}_j(\eta)$ and $\mathbf{u}^{\mathrm{ref}}_{r(j)}$ are the predicted and reference chunk directions. The no-regression submetric is
\begin{equation}
\label{eq:app_reg}
\tilde e_{\mathrm{reg}}(j, \eta) = \mathbf{1}\{\,\hat v_j(\eta) \ge \hat v_j(\theta_0) - \delta_{\mathrm{reg}}\,\},
\end{equation}
with geometric row score
\begin{equation}
\label{eq:app_vhat}
\hat v_j(\eta) = \frac{\sum_{m \in \mathcal{M}^{\mathrm{geom}}_j} \lambda_m\, \tilde e_m(j, \eta)}{\sum_{m \in \mathcal{M}^{\mathrm{geom}}_j} \lambda_m + \epsilon}
\end{equation}
where $\mathcal{M}^{\mathrm{geom}}_j$ contains the applicable geometric submetrics. The coefficients $\lambda_m$ and margin $\delta_{\mathrm{reg}}$ are fixed before ranking and shared across candidates. The coefficients are nonnegative and renormalized over the metrics applicable to each row.

\subsection{Wrong-target, invalid-pose, and unsafe blocker}
\label{app:blocker}
The validation blocker is
\begin{equation}
\label{eq:app_bj}
B_j(\eta) = \mathbf{1}\{\,\neg\,\mathrm{wrong\_target}(j, \eta) \ \wedge\ \neg\,\mathrm{unsafe}(j, \eta)\,\}.
\end{equation}
Here \emph{wrong target} includes entry of any predicted TCP into a declared non-target region or an endpoint orientation error exceeding the shared 0.35 rad tolerance. Non-target regions use point sets captured for the other components in the same configuration. The unsafe predicate flags workspace excursions, action-bound violations after inverse normalization, or an unsafe stop source. Terminal membership is scored separately, allowing valid partial approaches to receive geometric credit.

\subsection{Audits and candidate ranking}
\label{app:audits}
The pre-training audit checks split leakage, gate consistency on training rows, zero imitation weight on gate-failing rows, a finite loss denominator, the upper bound on row weights, and a non-collapsed effective weighted horizon count
\begin{equation}
\label{eq:n_eff}
n_{\mathrm{eff}}(\eta)\;=\;\frac{\bigl(\sum_{i\in\mathcal{D}_{\mathrm{tr}}}\sum_{h=1}^{H} w_i(\eta)\,m_{i,h}\bigr)^2}{\sum_{i\in\mathcal{D}_{\mathrm{tr}}}\sum_{h=1}^{H} \bigl(w_i(\eta)\,m_{i,h}\bigr)^2 + \epsilon}.
\end{equation}
The post-training audit requires zero wrong-target and unsafe predictions on the validation scoring and audit rows, plus per-component scoring-subset performance no lower than $\pi_{\theta_0}$ minus the no-regression margin fixed before ranking, under the same scoring protocol.

All eight candidates pass both audits. Table~\ref{tab:app_ranking} reports their ranking by $J_{\mathrm{val}}$.

\begin{table}[t]
\centering
\small
\caption{Candidate ranking by the component-balanced internal validation score $J_{\mathrm{val}}$, computed over the same fixed 294-row planner-phase validation subset for every candidate.}
\label{tab:app_ranking}
\begin{tabular}{lcc}
\toprule
Candidate & Internal validation score & Selected \\
\midrule
terminal\_stop\_heavy    & 0.770760 & yes \\
hover\_proximity\_heavy  & 0.770343 & no \\
correction\_heavy        & 0.769622 & no \\
rw\_like                 & 0.769558 & no \\
balanced\_low\_clip      & 0.769395 & no \\
conservative             & 0.769244 & no \\
ram\_connector\_recovery & 0.768812 & no \\
progress\_heavy          & 0.768420 & no \\
\bottomrule
\end{tabular}
\end{table}

The eight candidates in Table~\ref{tab:app_ranking} span less than $3 \times 10^{-3}$ in validation score. Validation resampling ranked the selected candidate first in 64.7\% of resamples. The two leading candidates together accounted for 94.6\% of top-ranked outcomes. Their score difference was 0.0004 with a 95\% interval [$-0.0011$,0.0024], so close candidates remained difficult to separate.

\section{Baseline Specifications}
\label{app:baselines}
The post-training procedures operate on the buffer $\mathcal{D}$. The initial demonstration set $\mathcal{D}_{\mathrm{demo}}$ is used only to train the shared starting checkpoint $\theta_0$, while the role \emph{clean} refers to supplementary demonstrations in $\mathcal{D}$. All baselines start from $\theta_0$ and use the same adapter recipe and hardware protocol, with the nominal update target given in Section~\ref{app:impl}.

PACER applies this budget to each of its eight predeclared candidates. Each post-training baseline fits one policy under a fixed weighting setting, while PACER uses finite validation selection. The budget fixes optimizer updates, not epochs, so smaller-data baselines receive repeated exposure to their included rows. Each procedure retains its declared parser filters, safety exclusions and per-horizon masks before loss computation. Table~\ref{tab:app_baselines} specifies each baseline, with the gate column indicating use of $g_i$ as a row-level inclusion rule. Outcome only uses supplementary clean rows and autonomous successes, identified by source or outcome labels and assigned uniform row weights.

Fixed geometry applies a precomputed rank-weight scalar to gate-passing rows, independently of the PACER candidate coefficients and validation selection.
\begin{equation}
\label{eq:app_fixed_geometry}
w_i^{\mathrm{FG}} = g_i\,\max\{0,\, r_i^{\mathrm{rank}}\},
\end{equation}
where $r_i^{\mathrm{rank}}$ combines progress, proximity, and target-region membership with fixed role multipliers and is fixed before PACER candidate selection.

The training data used by selected PACER exclude 227 autonomous rows flagged for safety-clamped actions. Uniform all-eligible and fixed geometry include those rows without that flag, giving 4613 positively weighted training rows versus 4386 for selected PACER (Table~\ref{tab:app_baselines}).

\begin{table}[t]
\centering
\footnotesize
\setlength{\tabcolsep}{4pt}
\caption{Baseline specifications. The gate column indicates the use of the PACER eligibility gate $g_i$. The train column counts the number of positively weighted training rows, the horizons column sums the mask over those rows, and the non-correction rows column counts positively weighted training rows in the clean, auto\_success, or partial roles.}
\label{tab:app_baselines}
\begin{tabular*}{\linewidth}{@{\extracolsep{\fill}}p{0.17\linewidth}p{0.15\linewidth}cccccc@{}}
\toprule
Baseline & Rows used & Train & Horizons & \shortstack{Non-correction\\rows} & Weight & Gate & Selected \\
\midrule
$\pi_{\theta_0}$ no post-train & none & 0 & 0 & 0 & none & no & no \\
Vanilla post-SFT & clean, correction & 3713 & 37130 & 2337 & 1 & no & no \\
Uniform all-eligible & all eligible & 4613 & 46130 & 3237 & 1 & yes & no \\
Outcome only & clean, auto\_success & 2662 & 26620 & 2662 & 1 & yes & no \\
Clean only & supp.\ clean & 2337 & 23370 & 2337 & 1 & yes & no \\
Correction only & correction & 1376 & 13760 & 0 & 1 & yes & no \\
Fixed geometry & eligible, rank weight & 4613 & 46130 & 3237 & $w_i^{\mathrm{FG}}$ & yes & no \\
PACER selected & eligible, positive $w$ & 4386 & 43860 & 3010 & $w_i(\eta^\star)$ & yes & yes \\
\bottomrule
\end{tabular*}
\end{table}

\section{Training and Implementation}
\label{app:impl}
PACER inherits the policy class, action decoder, and per-step targets of the flow-matching action-chunk planner $\pi_{\theta_0}$ unchanged. Unless otherwise stated, minibatches are sampled uniformly from the training rows included by each method, with PACER weight magnitudes applied in the native action loss.

The backbone is the $\pi_{0.5}$ model checkpoint produced by the initial supervised fine-tuning stage, and every post-training run trains a new LoRA adapter from this shared checkpoint, with PaliGemma rank 16 and alpha 16, action-expert rank 32 and alpha 32, and no LoRA dropout.

The optimizer is AdamW with $\beta_1 = 0.9$, $\beta_2 = 0.95$, and $\epsilon = 10^{-8}$, a 2000-step warmup to a constant $2 \times 10^{-5}$ learning rate, weight decay $10^{-10}$, and gradient clipping at global norm 1.0. Each fit uses a batch size of 32 and a nominal target of 10000 optimizer updates on 2 A100 GPUs with 8 CPU cores and 256 GB of memory. The eight-candidate search uses one training seed per setting.

Control and inference run at 10 Hz. All local policy inference, for both data collection and the held-out evaluation, runs on a workstation with an Intel Core Ultra 9 285K CPU, an NVIDIA RTX 5090 GPU with 32 GB of VRAM, and 64 GB of RAM.

The policy outputs chunks of 7-D joint-space actions, each comprising six joint deltas and an absolute gripper command. These chunks are mapped to tool-center-point poses in the robot base frame for scoring. The chunk horizon is $H = 10$ steps with no separate extraction stride.

Inputs are the base and wrist RGB views resized to $224 \times 224$, normalized with the OpenPI preprocessing statistics associated with each training manifest. No augmentation is used beyond resize and standard input transforms. Loss reduction follows Eq.~\eqref{eq:loss} with a $10^{-8}$ denominator floor.

The nominal seed fits and perturbed fit in Section~\ref{app:rebuttal_sensitivity} retain their respective training-native normalization.

\section{Extended Results}
\label{app:results}

\subsection{Held-out success with confidence intervals}
\label{app:ci}
Table~\ref{tab:main_results} reports overall held-out success and Wilson 95\% intervals. Table~\ref{tab:app_diff} adds descriptive paired component-cluster bootstrap intervals over the five components.

\begin{table}[t]
\centering
\small
\caption{Paired success-rate differences with descriptive component-cluster bootstrap 95 percent intervals over the five components.}
\label{tab:app_diff}
\begin{tabular}{lcc}
\toprule
Comparison & Difference & 95\% CI (pp) \\
\midrule
PACER $-$ vanilla post-SFT     & 33 pp & [14, 49] \\
PACER $-$ uniform all-eligible & 23 pp & [9, 40] \\
PACER $-$ fixed geometry       & 8 pp  & [3, 12] \\
\bottomrule
\end{tabular}
\end{table}

\subsection{Per-component comparison}
\label{app:percomp}
\begin{figure}[t]
\centering
\includegraphics[width=\linewidth]{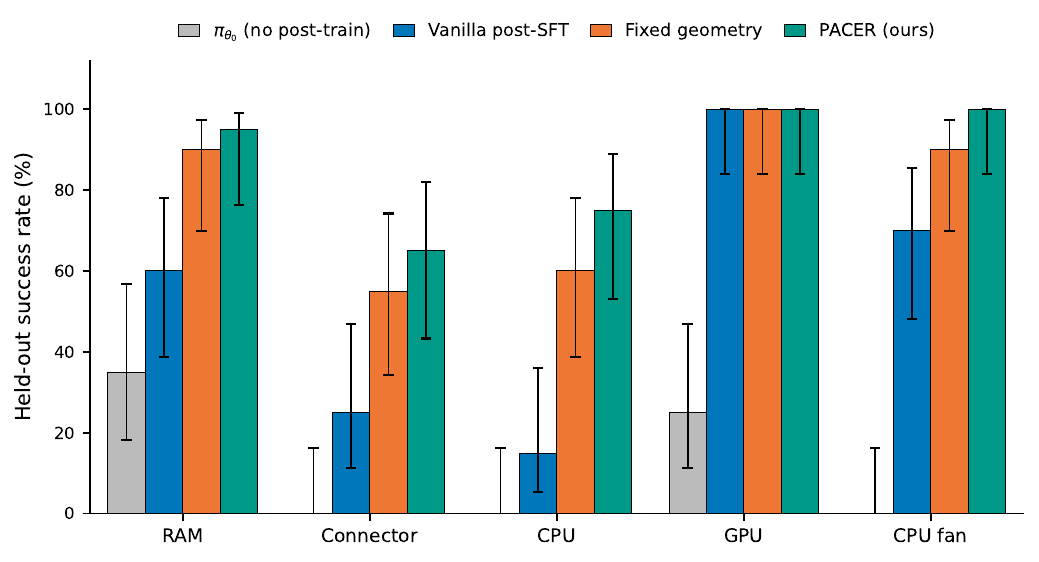}
\caption{Per-component held-out success rate over 20 trials per component. Error bars are 95\% Wilson binomial confidence intervals. PACER matches or exceeds fixed geometry on every component.}
\label{fig:app_percomp}
\end{figure}

Figure~\ref{fig:app_percomp} shows held-out success by component with 95\% Wilson intervals. PACER matches or exceeds fixed geometry on every component, with Connector and CPU among the components showing the largest margins. Each component has 20 trials per method, so the within-component intervals are wide.

Outcome only and the no-post-train reference have similar point estimates, at 11/100 and 12/100, respectively. GPU is at ceiling for several methods. On CPU fan, PACER reaches ceiling while all baselines remain below it.

\subsection{Hardware comparison of the eight PACER candidates}
\label{app:internal8}

In a separate internal hardware comparison, we evaluated one policy checkpoint for each of the eight weighting settings in Table~\ref{tab:app_candidates} across two hardware setups, denoted Setup~1 and Setup~2. Each candidate has 10 trials per setup and component, yielding 100 trials per candidate and 800 trials in total. Outcomes use the recorded operator-assigned binary success labels, with no trials excluded. Table~\ref{tab:app_internal8_overall} reports overall and setup-level success, while Tables~\ref{tab:app_internal8_components} and \ref{tab:app_internal8_setups} provide pooled and setup-specific component counts, respectively.

The highest observed overall success is 74/100 for terminal\_stop\_heavy, followed by 72/100 for progress\_heavy; both reach 30/50 on Setup~2. Their $+2$ percentage-point difference has a percentile 95\% interval of $[-6,11]$, obtained from 2,000 paired bootstrap resamples of recorded trial-index blocks within each setup--component stratum. The interval leaves the leading pair difficult to distinguish overall. Their component profiles differ: terminal\_stop\_heavy records more successes on Connector, CPU, and GPU, whereas progress\_heavy records more on CPU fan; they tie on RAM.

\begin{table}[t]
\centering
\footnotesize
\setlength{\tabcolsep}{3pt}
\renewcommand{\arraystretch}{1.08}
\caption{Overall and setup-level hardware success for the eight PACER candidates. Each candidate has 100 trials, with 50 per setup. Confidence intervals are marginal Wilson 95\% intervals for the overall success rate. Bold marks success-count column maxima.}
\label{tab:app_internal8_overall}
\begin{tabular*}{\linewidth}{@{\extracolsep{\fill}}lcccc@{}}
\toprule
Candidate & Overall & 95\% CI (\%) & Setup~1 & Setup~2 \\
\midrule
terminal\_stop\_heavy    & $\mathbf{74/100}$ & $[64.63, 81.60]$ & $\mathbf{44/50}$ & $\mathbf{30/50}$ \\
progress\_heavy          & $72/100$ & $[62.51, 79.86]$ & $42/50$ & $\mathbf{30/50}$ \\
conservative             & $66/100$ & $[56.28, 74.54]$ & $40/50$ & $26/50$ \\
ram\_connector\_recovery & $59/100$ & $[49.20, 68.13]$ & $39/50$ & $20/50$ \\
rw\_like                 & $59/100$ & $[49.20, 68.13]$ & $36/50$ & $23/50$ \\
hover\_proximity\_heavy  & $55/100$ & $[45.24, 64.39]$ & $36/50$ & $19/50$ \\
correction\_heavy        & $53/100$ & $[43.29, 62.49]$ & $33/50$ & $20/50$ \\
balanced\_low\_clip      & $45/100$ & $[35.61, 54.76]$ & $32/50$ & $13/50$ \\
\bottomrule
\end{tabular*}
\end{table}

\begin{table}[t]
\centering
\footnotesize
\setlength{\tabcolsep}{3pt}
\renewcommand{\arraystretch}{1.08}
\caption{Success counts by component pooled across the two hardware setups. Each component has 20 trials per candidate, for 100 trials overall. Bold marks success-count column maxima.}
\label{tab:app_internal8_components}
\begin{tabular*}{\linewidth}{@{\extracolsep{\fill}}lcccccc@{}}
\toprule
Candidate & RAM & Connector & CPU & GPU & CPU fan & Overall \\
\midrule
terminal\_stop\_heavy    & $\mathbf{15/20}$ & $\mathbf{9/20}$ & $17/20$ & $\mathbf{20/20}$ & $13/20$ & $\mathbf{74/100}$ \\
progress\_heavy          & $\mathbf{15/20}$ & $6/20$ & $14/20$ & $18/20$ & $\mathbf{19/20}$ & $72/100$ \\
conservative             & $10/20$ & $6/20$ & $15/20$ & $19/20$ & $16/20$ & $66/100$ \\
ram\_connector\_recovery & $12/20$ & $4/20$ & $15/20$ & $13/20$ & $15/20$ & $59/100$ \\
rw\_like                 & $4/20$  & $2/20$ & $\mathbf{19/20}$ & $16/20$ & $18/20$ & $59/100$ \\
hover\_proximity\_heavy  & $3/20$  & $7/20$ & $\mathbf{19/20}$ & $12/20$ & $14/20$ & $55/100$ \\
correction\_heavy        & $8/20$  & $2/20$ & $17/20$ & $12/20$ & $14/20$ & $53/100$ \\
balanced\_low\_clip      & $10/20$ & $3/20$ & $15/20$ & $10/20$ & $7/20$  & $45/100$ \\
\bottomrule
\end{tabular*}
\end{table}

\begin{table}[t]
\centering
\footnotesize
\setlength{\tabcolsep}{3pt}
\renewcommand{\arraystretch}{1.08}
\caption{Success counts within each hardware setup. Component cells report successes out of 10 trials; each setup total is out of 50. Bold marks success-count column maxima within each setup.}
\label{tab:app_internal8_setups}
\begin{tabular*}{\linewidth}{@{\extracolsep{\fill}}lcccccc@{}}
\toprule
Candidate & RAM & Connector & CPU & GPU & CPU fan & Setup total \\
\midrule
\multicolumn{7}{l}{\textbf{(a) Setup~1}} \\
\addlinespace[2pt]
terminal\_stop\_heavy    & $5/10$ & $\mathbf{9/10}$ & $\mathbf{10/10}$ & $\mathbf{10/10}$ & $\mathbf{10/10}$ & $\mathbf{44/50}$ \\
progress\_heavy          & $\mathbf{7/10}$ & $6/10$ & $\mathbf{10/10}$ & $9/10$  & $\mathbf{10/10}$ & $42/50$ \\
conservative             & $4/10$ & $6/10$ & $\mathbf{10/10}$ & $\mathbf{10/10}$ & $\mathbf{10/10}$ & $40/50$ \\
ram\_connector\_recovery & $6/10$ & $3/10$ & $\mathbf{10/10}$ & $\mathbf{10/10}$ & $\mathbf{10/10}$ & $39/50$ \\
rw\_like                 & $4/10$ & $2/10$ & $\mathbf{10/10}$ & $\mathbf{10/10}$ & $\mathbf{10/10}$ & $36/50$ \\
hover\_proximity\_heavy  & $1/10$ & $7/10$ & $\mathbf{10/10}$ & $8/10$  & $\mathbf{10/10}$ & $36/50$ \\
correction\_heavy        & $2/10$ & $2/10$ & $\mathbf{10/10}$ & $9/10$  & $\mathbf{10/10}$ & $33/50$ \\
balanced\_low\_clip      & $4/10$ & $3/10$ & $\mathbf{10/10}$ & $\mathbf{10/10}$ & $5/10$  & $32/50$ \\
\midrule
\multicolumn{7}{l}{\textbf{(b) Setup~2}} \\
\addlinespace[2pt]
terminal\_stop\_heavy    & $\mathbf{10/10}$ & $0/10$ & $7/10$ & $\mathbf{10/10}$ & $3/10$ & $\mathbf{30/50}$ \\
progress\_heavy          & $8/10$  & $0/10$ & $4/10$ & $9/10$  & $\mathbf{9/10}$ & $\mathbf{30/50}$ \\
conservative             & $6/10$  & $0/10$ & $5/10$ & $9/10$  & $6/10$ & $26/50$ \\
ram\_connector\_recovery & $6/10$  & $\mathbf{1/10}$ & $5/10$ & $3/10$  & $5/10$ & $20/50$ \\
rw\_like                 & $0/10$  & $0/10$ & $\mathbf{9/10}$ & $6/10$  & $8/10$ & $23/50$ \\
hover\_proximity\_heavy  & $2/10$  & $0/10$ & $\mathbf{9/10}$ & $4/10$  & $4/10$ & $19/50$ \\
correction\_heavy        & $6/10$  & $0/10$ & $7/10$ & $3/10$  & $4/10$ & $20/50$ \\
balanced\_low\_clip      & $6/10$  & $0/10$ & $5/10$ & $0/10$  & $2/10$ & $13/50$ \\
\bottomrule
\end{tabular*}
\end{table}

\subsection{Seed and geometry sensitivity}
\label{app:rebuttal_sensitivity}
We evaluate four policy fits on the same configuration, using 50 trials per fit for 200 trials in total. This configuration differs from those in $\mathcal{T}_{\mathrm{test}}$ and the controlled OOD comparison. Table~\ref{tab:app_sensitivity} reports Wilson 95\% intervals for each fit.

The three selected-policy training seeds 42, 43 and 44 achieve 33/50, 35/50 and 32/50, respectively (mean 66.7\%, standard deviation 3.1 percentage points). A policy retrained after a 10\,mm target-region translation and 0.1\,rad rotation achieves 38/50. Its difference from the nominal seed 42 policy is +10 percentage points with interval [$-7.7$,26.9].

\begin{table}[!htbp]
\centering
\small
\caption{Seed and geometry sensitivity, with 50 hardware trials per fit and Wilson 95\% intervals.}
\label{tab:app_sensitivity}
\begin{tabular}{lcc}
\toprule
Training run & Hardware success & 95\% CI (\%) \\
\midrule
Selected policy, seed 42 & 33/50 & [52.2,77.6] \\
Selected policy, seed 43 & 35/50 & [56.2,80.9] \\
Selected policy, seed 44 & 32/50 & [50.1,75.9] \\
Geometry perturbation & 38/50 & [62.6,85.7] \\
\bottomrule
\end{tabular}
\end{table}

\subsection{Annotation consistency}
\label{app:annotation_consistency}
In three 100-case relabeling sets, two raters agreed on 100, 93 and 98 cases for success, near and wrong location, respectively. This supports consistency under the labeling rubric in Section~\ref{app:labels}.

\subsection{Qualitative rollout sequences}
\label{app:qualrollouts}
\begin{figure}[h]
\centering
\includegraphics[width=\linewidth]{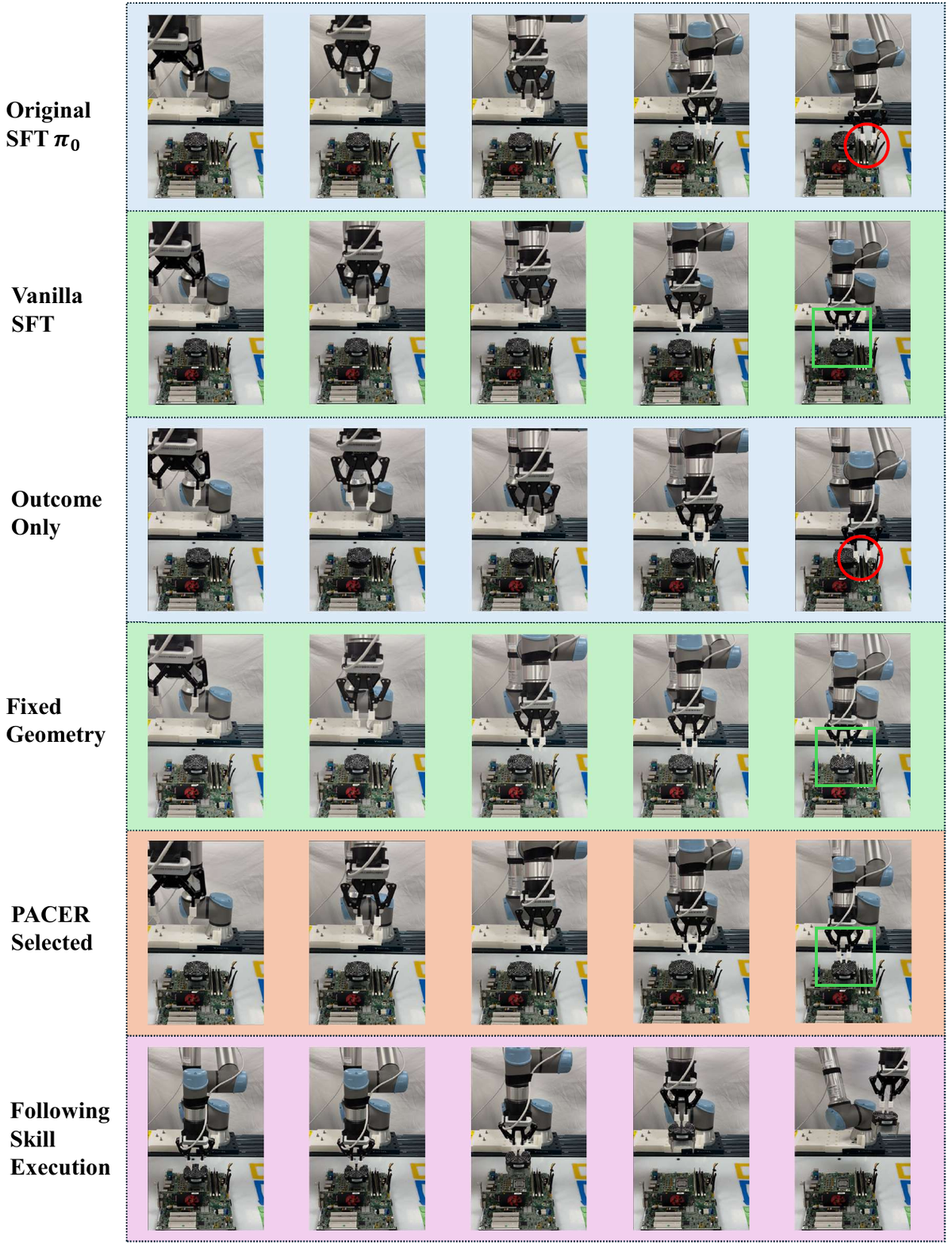}
\caption{Held-out comparison on the CPU fan task across five methods. The first five rows show the reference policy $\pi_{\theta_0}$, vanilla post-SFT, outcome-only, fixed geometry, and selected PACER, in that order. Red circles mark wrong-location or off-target endpoints, yellow triangles mark near-miss endpoints, and green boxes mark endpoints that reach the target region. The sixth row is a separate, unscored illustration of the fixed downstream skill executed from a reached handoff state.}
\label{fig:app_seq_cpufan}
\end{figure}

\begin{figure}[h]
\centering
\includegraphics[width=\linewidth]{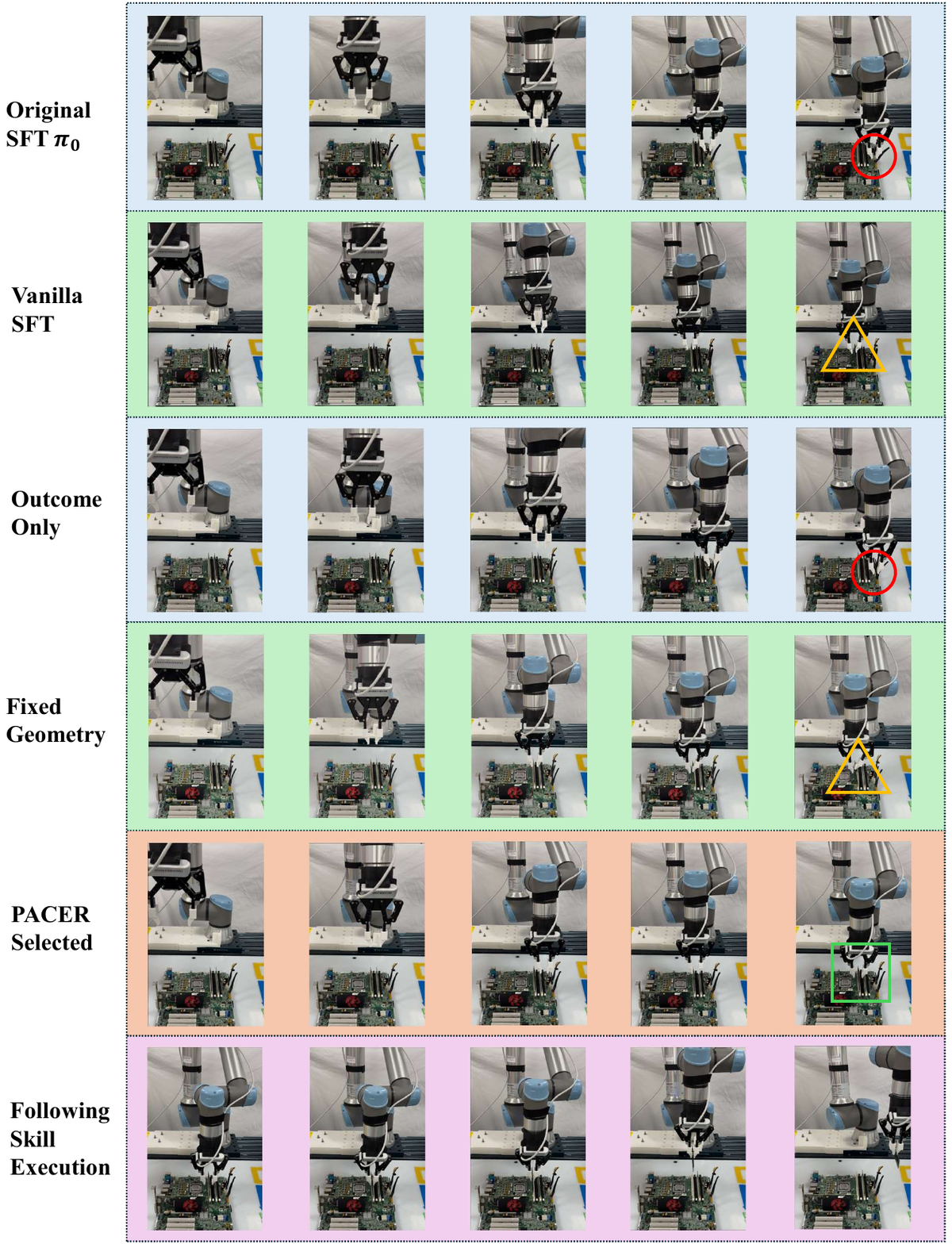}
\caption{Held-out comparison on the RAM task across five methods. Row order and endpoint markers are as in Figure~\ref{fig:app_seq_cpufan}.}
\label{fig:app_seq_ram}
\end{figure}

\begin{figure}[h]
\centering
\includegraphics[width=\linewidth]{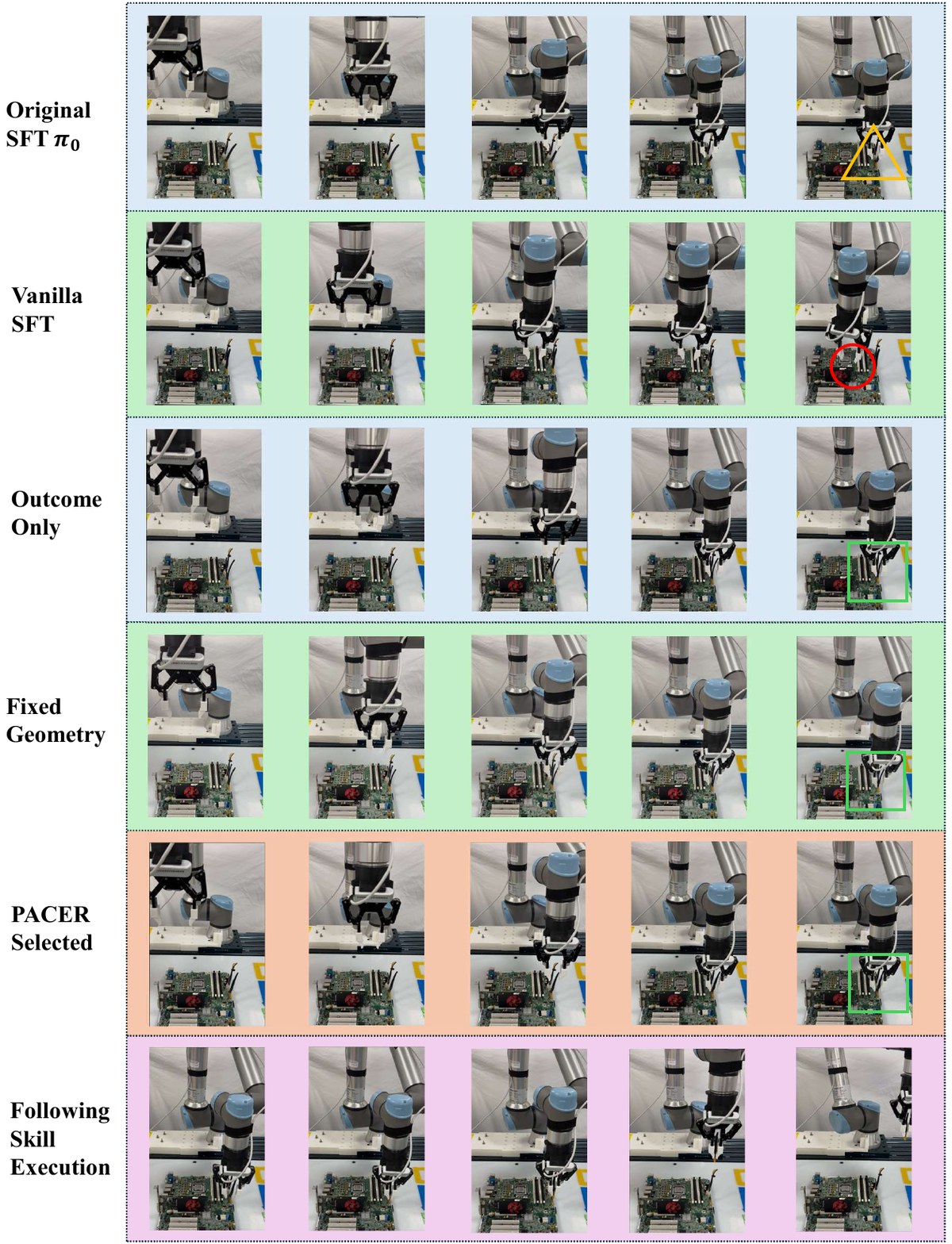}
\caption{Held-out comparison on the connector block task across five methods. Row order and endpoint markers are as in Figure~\ref{fig:app_seq_cpufan}.}
\label{fig:app_seq_connector}
\end{figure}

\begin{figure}[h]
\centering
\includegraphics[width=\linewidth]{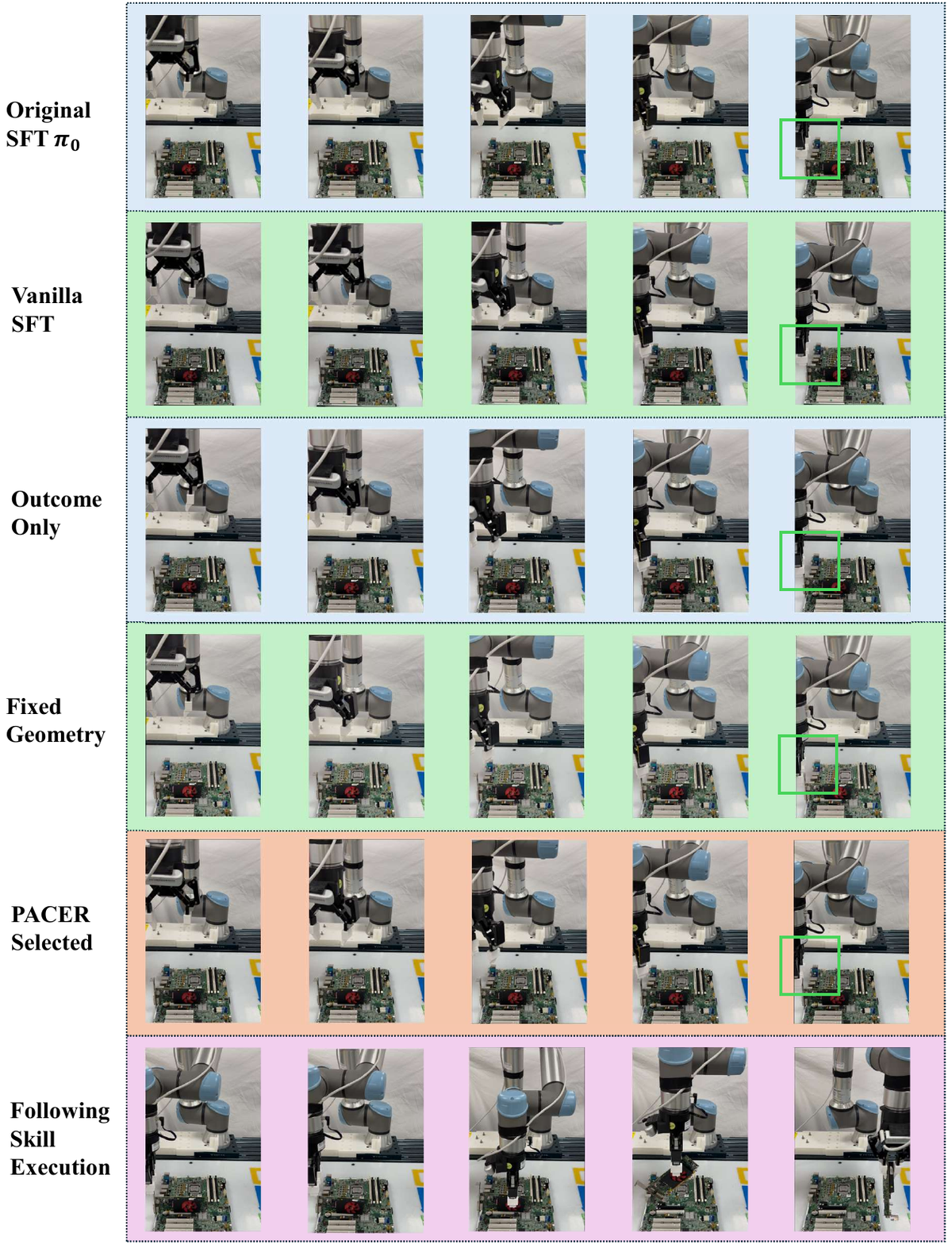}
\caption{Held-out comparison on the GPU task across five methods. Row order and endpoint markers are as in Figure~\ref{fig:app_seq_cpufan}.}
\label{fig:app_seq_gpu}
\end{figure}

\begin{figure}[tp]
\centering
\includegraphics[width=\linewidth]{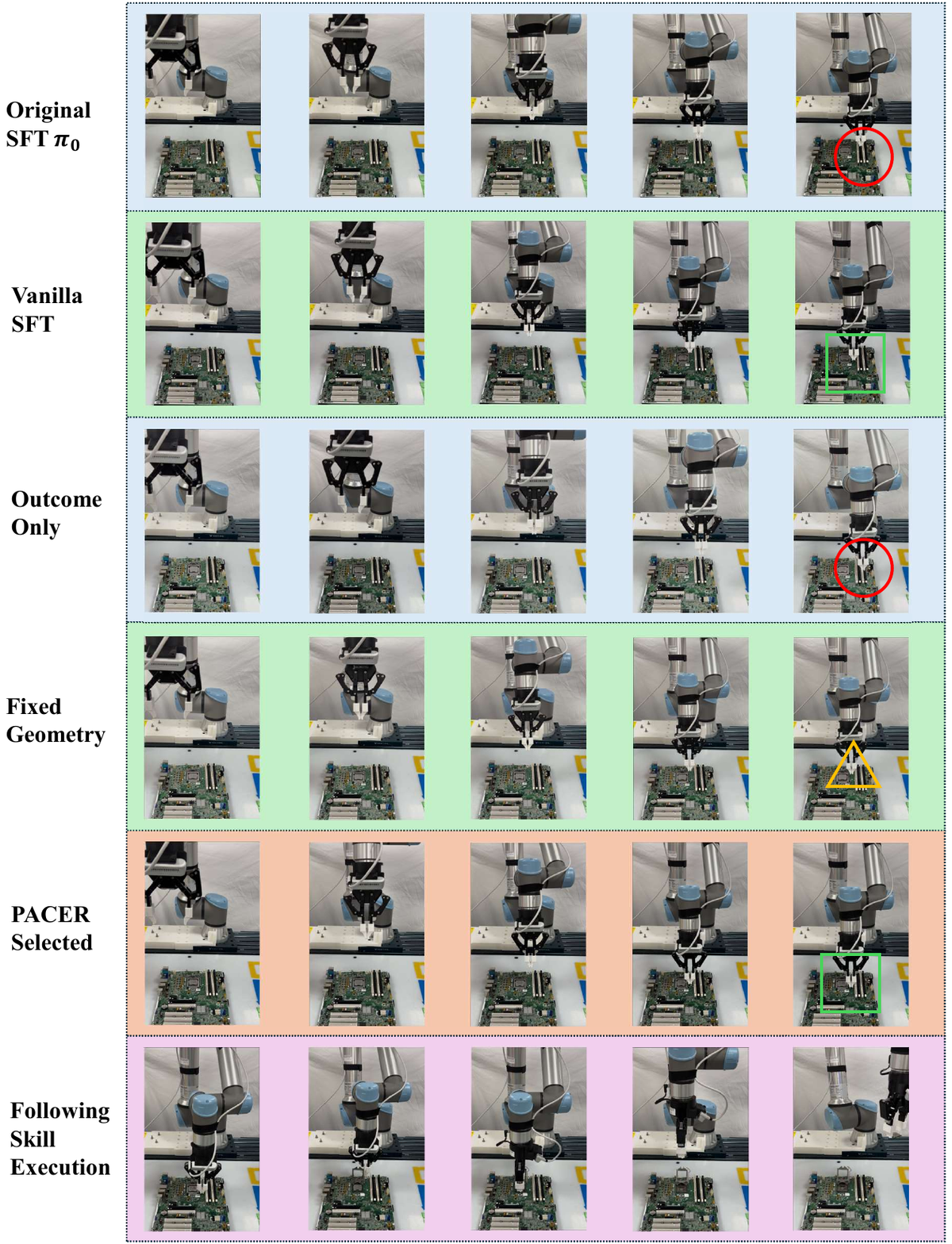}
\caption{Held-out comparison on the CPU task across five methods. Row order and endpoint markers are as in Figure~\ref{fig:app_seq_cpufan}.}
\label{fig:app_seq_cpu}
\end{figure}

Figures~\ref{fig:app_seq_cpufan} to \ref{fig:app_seq_cpu} show the five component tasks in the fixed evaluation order. Each figure has five method rows read left to right, with method order and endpoint markers defined in Figure~\ref{fig:app_seq_cpufan} and labels in Section~\ref{app:labels}.

\end{document}